\documentclass{article}

\usepackage[pdftex]{graphicx}
\usepackage[preprint,nonatbib]{neurips_2022}

\usepackage[utf8]{inputenc} 
\usepackage[T1]{fontenc}    

\usepackage{amsmath}
\usepackage{amsthm}
\usepackage{amssymb}

\usepackage{hyperref}       
\usepackage{cleveref}       
\usepackage{url}            
\usepackage{booktabs}       
\usepackage{amsfonts}       
\usepackage{nicefrac}       
\usepackage{microtype}      
\usepackage{xcolor}         

\usepackage{todonotes} 

\usepackage{pgfplots}
\pgfplotsset{width=10cm,compat=1.9}
\usepgfplotslibrary{external}
\usepackage{placeins}

\usepackage{multirow}

\usepackage{graphicx}

\usepackage{caption}
\usepackage{subcaption}
\usepackage{algorithm}
\usepackage{algpseudocode}

\newcommand{\thomas}[1]{\todo[inline,color=green!20!white]{\textbf{Thomas:} #1}}
\newcommand{\peter}[1]{\todo[inline,color=blue!20!white]{\textbf{Peter:} #1}}

\graphicspath{ {./images/} }

\newtheorem{lemma}{Lemma}

\DeclareMathOperator{\V}{Var}

\DeclareMathOperator{\E}{E}
\DeclareMathOperator{\Ent}{{\cal H}}
\DeclareMathOperator{\diag}{diag}

\newcommand{\R}{\mathbb{R}}
\newcommand{\eps}{\varepsilon}
\DeclareMathOperator{\jsd}{{\cal J}}
\DeclareMathOperator{\sm}{S}
\DeclareMathOperator{\tr}{Tr}

\newcommand{\SILENT}[1]{}

\newcommand{\bb}[1]{\mathbf{#1}}
\newcommand{\bbb}{\bb{b}}

\newcommand{\bbg}{\bb{g}}
\newcommand{\bbx}{\bb{x}}
\newcommand{\bbp}{\bb{p}}
\newcommand{\bbq}{\bb{q}}

\newcommand{\bbv}{\bb{v}}
\newcommand{\bbxin}{\bb{x}}
\newcommand{\bbxout}{\bb{y}}

\newcommand{\bby}{\bb{y}}
\newcommand{\bbw}{\bb{w}}
\newcommand{\bbz}{\bb{z}}
\newcommand{\bbdelta}{\boldsymbol{\delta}} 
\newcommand{\bbDelta}{\boldsymbol{\Delta}} 
\newcommand{\bbSigma}{\boldsymbol{\Sigma}} 

\newcommand{\SUP}[2]{{#1}^{({#2})}}
\newcommand{\inp}[2]{{\langle {#1}, {#2} \rangle}}

\newcommand{\Wmean}{\overline{W}}
\newcommand{\bbbmean}{\bar{\bbb}}
\newcommand{\bbmu}{{\boldsymbol\mu}} 

\newcommand{\SSigma}{\boldsymbol{\Sigma}}
\newcommand{\bbmuin}{\bbmu}
\newcommand{\bbmuout}{\bbmu'}
\newcommand{\Sigmain}{\SSigma}
\newcommand{\Sigmaout}{\SSigma'}

\newcommand{\act}{{A}} 

\newcommand{\bydef}{\stackrel{\text{def}}{=}}

\DeclareMathOperator{\hard}{\odot}
\DeclareMathOperator{\kprod}{\otimes}
\DeclareMathOperator{\argmin}{argmin}

\title{Favour: FAst Variance Operator \\for Uncertainty Rating}

\author{%
  Thomas D. Ahle \\
  Meta\\
  \texttt{ahle@fb.com} \\
  \And
  Sahar Karimi \\
  Meta \\
  \texttt{sahark@fb.com} \\
  \AND
  Peter Tak Peter Tang \\
  Meta \\
  \texttt{ptpt@fb.com} \\
}

\begin{document}

\maketitle

\begin{abstract}
Bayesian Neural Networks (BNN) have emerged as a crucial approach for
interpreting ML predictions.
By sampling from the posterior distribution, data scientists may estimate the uncertainty of an inference.
Unfortunately many inference samples are often needed, the overhead of which greatly hinder BNN's wide adoption. To mitigate this, previous work proposed propagating the first and second moments of the posterior directly through the network.
However, on its own this method is even slower than sampling, so the propagated variance needs to be approximated such as assuming independence between neural nodes. The resulting trade-off between quality and inference time did not match even plain Monte Carlo sampling.


Our contribution is a more principled variance propagation framework based on ``spiked covariance matrices'', 
which smoothly interpolates between quality and inference time.
This is made possible by a new fast algorithm for updating a diagonal-plus-low-rank matrix approximation under various operations.
We tested our algorithm against sampling based MC Dropout and 
Variational Inference on a number of downstream uncertainty themed tasks, 
such as calibration 
and out-of-distribution testing. We find that Favour is as fast as 
performing 2-3 inference samples, while matching the performance 
of 10-100 samples. 

In summary, this work enables the use of BNN in the realm of performance critical tasks where 
they have previously been out of reach.
\end{abstract}

\section{Introduction}

Machine learning and Neural Network (NN) in particular have greatly
exceeded expectations in their effectiveness at a diverse set of workloads
including classifications, recommendations, anomaly detection, human-machine
interaction, autonomous navigation; and the list goes on. Traditional NN
is learned (trained) by optimizing a loss function with a set of network 
parameters (weights) and input data as the function's arguments. 
The weights are real numbers and the learned NN produces point estimates
of the answer to the task at hand. For example, a binary classification
NN may render a judgement that the medical image examined shows that
there exist a 70\% chance of cancer. But the traditional NN cannot convey the
uncertainty of the opinion of 70\%. Nevertheless, answers with some
quantification of uncertainty are crucial as many critical and previously
exclusively human controlled tasks are increasingly shared with artificial
intelligence. For this example, it makes a big difference whether
the chance of cancer is $70\pm 2\%$ or $70\pm 25\%$.
The use for uncertainty
quantification in ML is evident in many auto decision processes.

\begin{figure}
    \centering
    \includegraphics[width=\textwidth]{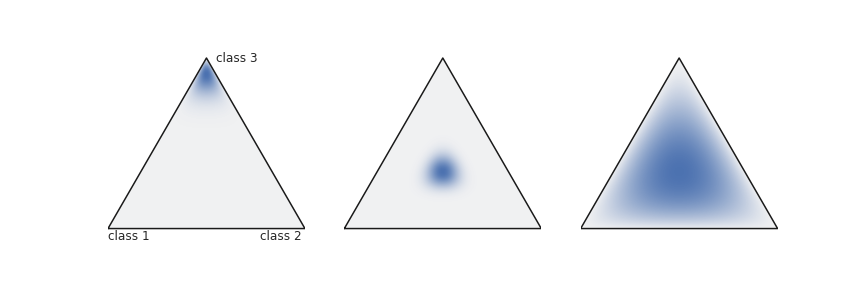}
    \caption{
    \textbf{Uncertainty in a classification problem with 3 classes.}
    Typical softmax inference gives a discrete probability distribution $(p_1, p_2, p_3)$, but in Bayesian NNs we predict a probability distribution over such probability distributions.
    In the picture:
    (a) A confident prediction of a class.
    (b) A confident prediction of data/aleatoric uncertainty.
    (c) High epistemic uncertainty, typical of out-of-distribution data.}
    \label{fig:distribution_as_result}
\end{figure}

In contrast to traditional NN, Bayesian Neural Network (BNN) treats the
network's weight parameters not as real numbers but real-valued 
random variables whose distributions are to be learned via training. Similarly, results of an inference made with BNN is not a real number but the distribution
of a real-valued random variable. 
See Figure~\ref{fig:distribution_as_result} for
the use of prediction distribution to reveal aleatoric and 
epistemic uncertainty.

Designing and training BNNs is an involved subject as the Bayesian framework
often involves intractable integration of probability densities. Innovative 
approximations that make training BNN practical constitute the bulk of active
researches. In comparison, attention to performing BNN inference is lacking.
One reason is that the conceptually easy method of sampling is applicable to many
(though not all) BNNs. After all, the weights are distributions. One can make
multiple inferences (take multiple samples) each on the same input data but with
different random drawing of the weights according to their distributions. 
Uncertainty quantification can be made with the collection of the
resulting point estimate results.
While sampling is easy to understand and implement, a moderate number
of samples in the order of tens, if not more, need to be collected to 
ensure an acceptable
uncertainty quantification. This consumes roughly ten fold of compute resources.
In many industrial scale problem where deterministic inferences fully utilize
compute capacity, this ten fold increase cannot be hidden at all.

Sample free methods for BNN inference have been proposed previously.
In the ``Expectation propagation'' framework we approximate the true posterior distribution with a simpler approximation.
Choosing a Multivariate Gaussian, minimizing the KL-Divergence to the true posterior is possible if we can compute the first two moments.
This means we follow the evolution of not just the input
multi-dimensional random variable when it passes through a BNN, but also its mean and covariance. Since the distribution of the involved
random variables are very often well characterised by these two
statistics, variance propagation can provide information comparable
to that obtained by large sampling. The main drawback is the
high cost of this technique. Roughly speaking, a linear layer
with a $n$-by-$n$ weight matrix has an inference cost of $O(n^2)$
while propagating the covariance costs $O(n^3)$. 

By approximating covariance matrices in a 
diagonal-plus-low-rank (DPLR) form, also called spiked covariance, 
we derived the evolution of these approximate covariance and
devised efficient algorithms to follow their evolution. 
Accuracy can improve with
larger rank; although our experiments show that a small rank of 2 
produces uncertainty rating of the quality comparable to that
obtained by large samplings.
The cost is of the order of performing inference with a deterministic
NN, and in the case of MLP, the cost can be made sub-inference through
further approximations. 

Our contributions are several fast algorithms that propagate
spiked variance through a general Bayesian Neural Network
layer thus covering many network models including regression,
computer vision and recommendation systems. Propagation of
the spiked covariance leads to sample free substitute for 
MC Dropout and Variational Inference. 
These fast algorithms
offer a Pareto curve of tradeoff between speed and accuracy, resulting in better uncertainty quantification then large samplings
at a lower cost.

In what follows, Section~\ref{sec:background} reviews 
related works. Section~\ref{sec:var-prop} presents the basic 
mathematics of mean and variance propagation. Section~\ref{sec:fast-prop} 
explains the computational algorithm we developed to propagate spiked
variance at inference complexity.
We also present other fast algorithms such as an approximation
to the Jensen Shannon Divergence.
Section~\ref{sec:experiments} presents results on using variance
propagation on a number of uncertainty rating tasks.
Section~\ref{sec:conclusion} summarises our results and present directions
for future work.

\section{Background}\label{sec:background}

Different designs for probabilistic neural networks have been made and
\cite{BNNTutorial_2020, Goan_2020, Mackay1995} give excellent survey
and tutorial on the subject. Some architectures most relevant to our work include the following.
A deep ensemble of models is suggested in \cite{deepEnsembles_2017} for attaining the predictive distribution. Despite its  simplicity, deep ensemble requires training multiple copies of the model which can be prohibitively expensive for realistic and large scale models. Batch ensemble \cite{batchEnsemble_2020} and Monte Carlo dropout \cite{mcDropout_2016} are scalable alternatives to formulating a model's weight
parameters as general distributions. Another class of scalable approaches is \emph{variational inference} (VI) \cite{Hinton1993KeepingTN}. Gaussian mean-field VI are proposed in \cite{VIforNN_2011, BBB_2015}. In \cite{rank1BNN}, Bayesian neural nets with rank-1 variational factor are suggested where weight matrices are Hadamard products of a deterministic matrix and a rank-1 variational one. Going beyond mean-field VI is investigated in \cite{SLANG, roubl}. We refer to these different architectures of probabilistic networks by the general term Bayesian Neural Network, or BNN. 
Training BNNs are more involved both conceptually as well as computationally than their traditional 
counter parts. Nevertheless, in situations where the input data distribution changes very
slowly, such as natural language processing or medical diagnosis, 
the cost of training
can be amortized over the lifetime of the network's deployment. But the
additional cost in making inference with BNN compared to traditional NN cannot be amortized in general.

In Bayesian inference, the predictive distribution of an unknown target 
$\bby$ is 
$p(\tilde \bby|\tilde \bbx,D) = \int_\bbw p(\tilde \bby|\tilde \bbx,\bbw) p(\bbw|D)$; in practice, 
however, most algorithms rely on sampling from the posterior of $w$ to 
attain the predictive posterior. MC Dropout~\cite{mcDropout_2016} is an alternative to sampling the posterior of $\bbw$. Here the weights
are deterministic, that is, real values; but probabilistic dropout
layers are used even during inference time. A number of inference
result samples are collected on one fixed input data. The samples
of inference (point estimate) results represent the posterior 
distribution of the BNN inference. Sampling is thus a convenient
approach to realize BNN inference for several classes of BNN designs.
Unfortunately, it is unsure how many samples one need to get a reliable representation of a posterior distribution. Empirically, ten or twenty samples
are used, and sometimes five samples are used in the face of
compute performance constraint. This constraint is real in many different situations: decision in autonomous navigation, recommendation systems
that need to remain highly responsive, real-time blockage of harmful contents, etc. 

Sample free methods have been proposed before such as
\cite{distDistillation_2021} where
a deterministic model that produces uncertainty
quantification is trained after the BNN in question is
trained. Alternatively, our approach closely resembles that of
\cite{varProp_2019}. It
addresses BNN expressed by Monte Carlo Dropout \cite{mcDropout_2016}.
The main idea is the mean and variance of a multidimensional random
variable $\bbx \in \R^n$ whose mean $\bbmu$ and variance $\boldsymbol{\Sigma}$
are known (perhaps approximately) will evolve deterministically.
In the MC Dropout setting, a linear layer $W\bbx + \bbb$ is deterministic
and thus the mean and variance after this layer becomes $W\bbmu + \bbb$
and $W\boldsymbol{\Sigma}W^T$, respectively. Unfortunately this propagation
has a higher order complexity compare to simple inference: If $W\in\R^{n\times n}$, then inference costs $O(n^2)$ but computing $W\boldsymbol{\Sigma}W^T$
costs $O(n^3)$. The straightforward complexity reduction by assuming
$\boldsymbol{\Sigma}$ to be diagonal, hence assuming the components
of $\bbx$ to be mutually independent, is too restrictive to offer any
gains over straightforward sampling.

We reduce the complexity of sample free variance propagation method
back to that of deterministic inference without assuming $\boldsymbol{\Sigma}$ 
to be diagonal, but of the general form of a non-negative diagonal
plus a low-rank semidefinite matrix. One can view our as an extension of \cite{Barber98ensemblelearning} unto inference. This method
is applicable to any BNN expressible in terms of
weight parameters or layers that are adequately determined by means
and variances. In particular, MC Dropout, or ``learned'' dropout
with per-neuron dropout distribution, mean-field and
generalized mean-field Variational Inference.

\SILENT{
\begin{figure}
    \hspace{-2.4cm}
    \resizebox{1.3\textwidth}{!}{
    \input{images/histogram.pgf}
    }
    \caption{
    \textbf{Variance propagation reduces inference noise.}
    In the figure an MLP was trained with dropout ($p=0.5$) on a noisy Sine dataset.
    Left: Monte Carlo Dropout inference with 10 samples.
    Right: Favour inference with co-variance rank 10, using the same dropout-trained model.
    }
    \label{fig:sine}
\end{figure}
}

\SILENT{
sampling free algorithms for improved inference: \cite{distDistillation_2021, varProp_2019}

Common tasks for uncertainty systems include
\begin{enumerate}
    \item Out of sample detection 
    
    \item Calibration (Brier/NLL/ece)
    \item Calibration (Regression confidence intervals / Coverage)
    \item Monotonicity under pertubation
    \item Transfer learning
\end{enumerate}

\peter{
Are these the only common tasks? And do we plan to show results in all of these tasks in that var prop yield better result than sampling.
}

\thomas{
I'm thinking we can show space/time trade-off cruves for each of those tasks. Except transfer learning, I'm not quite sure what that one is.
}

\subsection{Contribution}
\begin{enumerate}
    \item Faster the previous Variance and Prop work
    
\end{enumerate}
https://www.overleaf.com/project/624f25131a29f203f4b6975d
}

\section{Mean and Variance Propagation}\label{sec:var-prop}

Our focus is on inference with a probabilistic network so as to produce 
result with some uncertainty rating. Sampling is a straightforward 
technique where inference with one sample of the probabilistic network 
and input produces one sample inference result. Gathering sufficient
result samples allows various uncertainty quantification be made.
Empirically, the number of samples $S$ needed to make 
reliable quantification is in the order of tens, which consumes $S$ times as much compute
resources.

A sample free alternative is this. Given a random variable $\bbxin\in\R^n$ 
with mean and (co)variance $(\bbmuin, \Sigmain)$, we wish to determine the
mean and variance $(\bbmuout,\Sigmaout)$ of the output $\bbxin\in\R^m$ 
of a generic probabilistic neural network layer of the
form
$\bbxout = \act(W(D\bbxin) + \bbb)$.
$\act(\cdot)$ is a deterministic non-linear activation function such 
as ReLU, sigmoid or tanh. The affine transformation $W\bbx + \bbb$ can 
be deterministic, that is, the entries in the $W$ matrix and $\bbb$ vector
are simple real numbers, or probabilistic when those entries are 
distributions. $D$ is a elementwise (diagonal matrix) generalized
dropout operator where each element is a scaled Bernoulli 
random variable of possibly different dropout probabilities $p$. 
That is, the value is 0 with probability $p$ and $1/q$ with probability 
$q=1-p$. The parameters $(\bbmu,\boldsymbol{\Sigma})$ characterize 
important distributions such as Gaussian that are considered good representation
of the underlying distributions of $\bbxout$. 
This generic layer encapsulates a wide range of 
network architectures and represents multiple Bayesian realizations
including MC Dropout and Variational Inference. 

Consider the simple case of $\bbxout = W \bbxin + \bbb$ 
where $W, \bbb$ are deterministic
but $\bbxin \sim (\bbmu,\Sigmain)$. Straightforward statistical derivations
show $\bbmuout = W\bbmuin + \bbb$ and
$\Sigmaout = W\,\Sigmain\,W^T$. 
In this section we state $(\bbmuout,\Sigmaout)$ for our generic BNN layer
and full derivations are given in the appendix for completeness.
We note that our treatment of 
affine transformation and non-linear functions are
more general than~\cite{varProp_2019} where the weights are
deterministic, \cite{HernndezLobato2015ProbabilisticBF} where
the neurons are assumed independent or~\cite{Mae2021} where one 
propagates a bound on the variance. We devote the next section to
the details where we reduce the apparent $O(n^3)$ complexity
of $\Sigmaout = W\,\Sigmain\,W^T$ to $O(n^2)$ or less.

\subsection{Propagation through Dropout Layers}\label{subsec:dropout}
Let $\bbp\in\R^n_+$ be the dropout probabilities for each of $\bbxin$'s components. 
$D$ is a scaled Bernoulli random variable such that $E[D] = I$, the identity matrix
yielding $\bbmuout = \bbmuin$. Then
$V[D\bbxin] = \Sigmain + \diag((\bbmuin^2 + \diag(\Sigmain))\hard \bbp/\bbq)$,
where $\hard$ is the Hadamard (pointwise) product.
The computational complexity is a low $O(n)$ and note that $\bbp/\bbq$ simplifies 
trivially to the scalar $p/q$ if dropout probabilities are identical for all of $\bbxin$'s components.

\subsection{Propagation through Linear Layers}
Consider now a linear layer $\bbxout = W \bbxin + \bbb$ where
$W \in \R^{m\times n}$, $\bbb\in\R^m$. Both $W$ and $\bbb$ can be 
distributions instead of deterministic, with $\Wmean$ and $\bbbmean$
be their respective means. Furthermore, the entries within $W$ or 
$\bbb$ need not be mutually independent while we always assume $W$ and 
$\bbb$ are independent of each other. We propagate the mean
$\bbmuout$ of $\bbxout$ by 
$\bbmuout = \Wmean\bbmuin + \bbbmean$. For the variance, we have
\begin{equation}\label{eqn:linear_general}
(\V[W\bbxin + \bbb])_{ij} = (\Wmean\Sigmain\Wmean^T + \Sigma_{\bbb})_{ij} + 
{\rm SUM}(\Sigma_{\bbw,ij} \hard (\Sigmain + \bbmuin \bbmuin^T))
\end{equation}

Equation~\ref{eqn:linear_general} simplifies in some common special cases.
\begin{enumerate}
    \item If the linear layer is deterministic, then $\Sigmain_{\bbw},\Sigmain_{\bbb}$ are zero and the variance of $\bbxout$ is simply $W\Sigmain W^T$.
    \item If the entries of $W$ and $\bbb$ are all mutually independent, such as
    in the case of mean-field VI, then $\Sigmain_{\bbb}, \Sigmain_{\bbw}$ are diagonal
    matrices. A natural way to represent $\Sigmain_{\bbw}$
    is by $W_{\sigma^2} \in \R^{m\times n}$ whose
    entries are the variances of $w_{ij}$ and for $\Sigmain_{\bbb}$ 
    the vector $\bbb_{\sigma^2}\in\R^m$. Thus $\Sigmaout =
    \Wmean \Sigmain \Wmean^T + \diag\left(
    \bbb_{\sigma^2} +
    W_{\sigma^2}\,(\diag(\Sigmain) + \bbmuin^2))
    \right).
    $
    \item Consider the generalized mean-field VI where each different rows of the $W$ matrix
    are independent, but not so withing each row. In this case, 
    ${\rm SUM}(\Sigma_{\bbw,ij} \hard (\Sigmain + \bbmuin \bbmuin^T))$ is non-zero
    only when $i=j$ and thus this expression reduces to
    ${\rm SUM}(\Sigma_{\bbw,ii} \hard (\Sigmain + \bbmuin \bbmuin^T))$.
\end{enumerate}

\subsection{Propagation through Univariate Activations}
An activation function $\act(\cdot)$ such as ReLU or sigmoid are univariate nonlinear functions
applied to each component of an in $\bbxin =[x_1, x_2, \ldots, x_n]^T$. 
Similar to existing work, we approximate the $i$-th component of $\bbmuout$
by taken the expectation of $A(x)$ using $x \sim  N(\mu_i,\sigma_i^2)$. This
computation of $\bbmuout$ is of $O(n)$ complexity.

There are two choices for variance propagation. The first approach uses the
first order Taylor expansion at the input mean $\bbmuin$:
$
\bbxout \approx \act(\bbmuin) + D\cdot(\bbxin-\bbmuin), \quad
D \bydef \diag([\act'(\bbmu_i); i=1, 2,\ldots, n]).
$
This leads to the approximation $\Sigmaout \approx D\Sigmain D^T$, which is merely
scaling $\Sigmain_{ij}$ by $\act'(\bbmuin_i)\act'(\bbmuin_j)$. In particular, for
ReLU activation, this is zeroing out the $i$-th row and column whenever $\bbmuin_i \le 0$. This sparsification may have have some desirable regularization effect.
Note however that the covariance values in $\Sigmain$ remain unchanged when not made zero. This does not match the case if we assume $\bbxin$ is Gaussian.
This motives our second approach. We compute
$\sigma'_i = \sqrt{\V[\act(x)]}$
for $x\sim N(\mu_i,\sigma_i^2)$ and
$D \bydef \diag([\sigma'_i/\sigma_i \text{if $\sigma_i > 0$ else 0}]$.
We scale $\Sigmain$ by
$D\Sigmain D^T$. This guarantees that the diagonal value
of $\Sigmaout$ matches the Gaussian assumption.

\section{Diagonal Plus Low Rank Approximation}\label{sec:fast-prop}

As described in the introduction, a Diagonal Plus Low Rank 
(DPLR) approximation to a covariance matrix is a natural way to reduce inference time while keeping a control on the quality trade-off incurred.
In the appendix (\cref{subsec:fast_common_operations}) we describe now many useful, established properties of this decomposition.
Here we present the most novel and interesting ideas for our application, which is propagating the approximation through the layers of a (Bayesian) neural network.

\SILENT{
now a method with inference complexity that approximate
$\Sigmaout$ in a diagonal-plus-low-rank (DPLR) form given a DLPR $\Sigmain$. 
As the very first $\Sigmain$ is most naturally in this form, when for example the
input features are deterministic or independent, we can propagate
variance from beginning to end with inference complexity. Furthermore,
for MLP models, we can further reduce the complexity to the sub-inference
regime. Given $\Sigmain$ in DPLR form, propagation through dropout and
activation layers described in the previous section preserves the DPLR
form with low complexity trivially. Thus in the section we focus on the Bayesian linear layer. We further develop several related computations on the DPLR
form.}

\subsection{Fast DPLR Linear Propagation }\label{subsec:fast_dplr_linear}

Consider the computation of $M = \Wmean \Sigmain \Wmean^T \in \R^{m\times m}$
where $\Wmean\in\R^{m\times n}$.
Straightforward computation requires $O(mn^2)$ operations.
We can however obtain a DPLR approximation to $M$ with complexity 
commensurate with model inference. We exploit the DPLR structure of
$\Sigmain = \Lambda + U U^T$ and use subspace iteration to obtain the DPLR approximation
to $M$ which only requires matrix-vector product with $M$ without needing $M$ explicitly.

We formulate our problem as
\[
\argmin_{\Lambda,V} \rho(\Lambda,V),\;
\Lambda=\diag(\boldsymbol{\lambda}),\boldsymbol{\lambda}\in\R_+^m,\; 
V\in\R^{m\times r}
\;\hbox{and}\;
\rho(\Lambda,V) \bydef \|M - (\Lambda+V V^T)\|_F^2,
\]
and give a heuristic algorithm that runs in $O(mnr + mr^2)$ time.
This is based on the alternating projections algorithm for Robust PCA algorithm~\cite{netrapalli2014non}, but we suspect the particular version we need is NP Hard to solve.

The basic idea is to update $V$ and $\Lambda$ alternately
while keeping the other fixed. 
Given the current
iterate $\Lambda_{\rm now}$, 
the symmetric matrix $M-\Lambda_{\rm now}$
has an eigendecomposition
\[
M-\Lambda_{\rm now} =
\psi_1 \bbz_1\bbz_1^T + \psi_2 \bbz_2 \bbz_2^T + \cdots
+ \psi_m \bbz_m \bbz_m^T
\]
$\psi_1 \ge \psi_2 \ge \cdots \ge \psi_m$.
The best Frobenius norm (as well as 2-norm)
rank-$r$ semi-definite approximation is
$\sum_{\ell=1}^k \psi^+_\ell \bbz_\ell \bbz_\ell^T$
where $\psi^+_\ell \bydef \max(\psi_\ell, 0)$
\cite{Higham1988}. Provided
$\psi_r > |\psi_j|$ for all $j > r$, 
the rank-$r$ truncated SVD of $M-\Lambda_{\rm now}$ is 
$Z_{(r)} \diag([\psi_1,\psi_2,\ldots,\psi_r])
Z^T_{(r)}$. In this case, we obtain 
$V_{\rm next} = Z_{(r)}\diag([\sqrt{\psi_1},\sqrt{\psi_2},\ldots,\sqrt{\psi_r}))$
Because this $V_{\rm next}$ is the best approximation, we have
$\rho(\Lambda_{\rm now},V_{\rm next})
\le \rho(\Lambda_{\rm now}, V_{\rm now})$.

On the other hand, fixing $V_{\rm now}$, it is easy
to see that
$\Lambda_{\rm next} \gets 
\max(\diag(M-V_{\rm now} V^T_{\rm now}), \mathbf{0})$
is the optimal and will lead to a descent of $\rho$.

Classical subspace iteration algorithms \cite{Mang1977} for 
obtaining $r$-truncated SVD to a matrix $M$
is well established and has order of complexity that of $M$ times $r$ vectors. 
Recent developments of randomized linear algebra~\cite{Gu2015} improve the performance further. 
We developed the following greedy algorithm by interleaving the alternating
direction search with a subspace iteration prompted by the effectiveness
of randomized initialization.

\begin{algorithm}
\caption{Fast DPLR Decomposition}
\label{algo:favour}
\begin{algorithmic}[1]
\Require $\Lambda, U$, $\Lambda \in\R_+^{n\times n}$ diagonal, $U\in\R^{n\times r}$, $W$; $M=W(\Lambda+UU^T)W^T$.
\State $V \sim N(0,1)^{m\times r}$
\Comment{Or best guess of final low-rank matrix, $O(mr)$ time}
\For{$i = 1, 2, \ldots, K$}
\Comment{Typically $K=2$, 3 or 4}
    \State $S \gets \diag([\|v_1\|_2, \|v_2\|_2,\ldots,\|v_r\|_2])$
    \Comment{$O(mr)$ time}
    \State $\Lambda\gets\diag(M - VS^{-1}V^T)$
    \Comment{$O(mr)$ time}
    \State $\Lambda\gets\max(\Lambda, 0)$
    \Comment{(Component-wise $\max$), $O(mr)$ time}
    \State $V \gets V S^{-1}$
    \label{alg-line:u-norm}
    \Comment{(Make columns unit-norm), $O(mr)$ time}
    \State $V \gets (M - \Lambda)V$
    \label{alg-line:matvec}
    \Comment{$O(mnr)$ time}
    \State $V \gets \hbox{Unnormalized-Gram-Schmidt$(V)$}$, 
    \Comment{$O(mr^2)$ time}
\EndFor
\State $S \gets \diag([\|v_1\|_2, \|v_2\|_2,\ldots,\|v_r\|_2])$
\Comment{$O(mr)$ time}
\State $\Lambda\gets\max(\diag(M - VS^{-1}V^T), 0)$
\Comment{$O(mr)$ time}
\State \Return $(\Lambda, V S^{-1/2})$
\end{algorithmic}
\end{algorithm}
The matrix-vectors product on Line \ref{alg-line:matvec} does not compute $M$ explicitly but
exploits the structure $M=W(\Lambda + UU^T)W^T$. See Section~\ref{subsec:fast_common_operations} for
more details.
The normalization of $U$ on Line \ref{alg-line:u-norm} is not strictly necessary, but if the number of loop iterations ($K$) is large we might otherwise get numerical overflows.
The dominant complexity is $O(mnr)$ which is of the order of
making $r$ inferences. In practice, $r$ is typically set between 2 and 4. Moreover, in the case of linear layers
where $W$ is dense and compute intensive, we have replaced
it with a low-rank approximation of rank $q \le n/10$ in
Algorithm~\ref{algo:favour} and obtained good uncertainty
rating result. In this case, the complexity is 
$O(mqr)$ which can be lower than model inference.

\subsection{Fast DPLR Propagation for Convolution}\label{subsec:fast_dlpr_conv}
Convolution is a linear operator but with special structure: Expressed
as a matrix it is sparse and Toeplitz. Nevertheless, we use
the convolution and transposed convolution operators to realize the
computation in the previous section by exploiting the DPLR structure.
The diagonal and each rank-1 component of the DPLR representation and
each ``vector'' used in subspace iteration is naturally in 2D shape. The
$W$ and $W^T$ operations translates to application of {\tt Conv2d}
and {\tt ConvTranspose2d} (in PyTorch for example).

\subsection{Fast Jensen-Shannon Divergence Approximation of DPLR Variance}
\label{subsec:fast_dplr_jsd}

\begin{figure}
     \centering
     \begin{subfigure}[b]{0.49\textwidth}
         \centering
         \includegraphics[width=\textwidth]{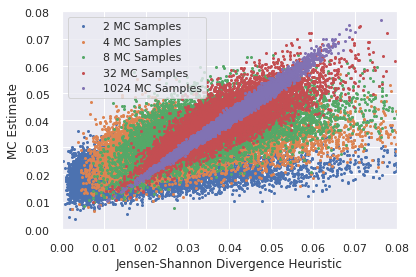}
         \caption{JSD on Softmax on 10 dimensional random Gaussian data with small covariance.}
         \label{fig:heur_small}
     \end{subfigure}
     \hfill
     \begin{subfigure}[b]{0.49\textwidth}
         \centering
         \includegraphics[width=\textwidth]{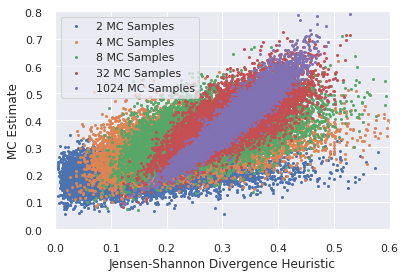}
         \caption{JSD on Softmax on 10 dimensional random Gaussian data with larger covariance.}
         \label{fig:heur_large_small}
     \end{subfigure}
    \caption{
    \textbf{Concordance between JSD estimates from samples and \cref{lemma:jsd-details}.}
    The more MC Dropout samples are used, the better the JSD formula agrees with the estimate.
    This suggests the approximation gets close to the true JSD much faster than the sampling method, specially when the JSD scores are not too large.}
    \label{fig:jsd}
\end{figure}

For classification problems, the most common last layer of the network
is a softmax operator. Typically, one collect samples of the softmax results
$\SUP{\bbp}{1},\SUP{\bbp}{2},\ldots,
\SUP{\bbp}{n}$ and computes the Jensen-Shannon Divergence
$
\jsd(
\SUP{\bbp}{1},\SUP{\bbp}{2},\ldots,
\SUP{\bbp}{n}
) =
\Ent(\overline{\bbp}) - \frac{1}{n}\sum_\ell\Ent(
\SUP{\bbp}{\ell})
$
where $\Ent(\bbp) \bydef \inp{-\bbp}{\log(\bbp)}$ is
the entropy. We proved that the following sample free
estimate accurately approximates the JSD:
$
\left(1-\frac{1}{n}\right)\frac{1}{2}
\left( \inp{\bbp^*}{\diag\bbSigma} - 
\inp{\bbp^*}{\bbSigma\bbp} 
\right)
$
where $\bbp^*=\sm(\bbmuin)$ is the softmax of the
mean $\bbmuin$. This computation is of $O(n)$ when
$\Sigmain$ is of DPLR form. The full statement
and proof are given in the Appendix.
Figure~\ref{fig:jsd} illustrates the estimate's effectiveness. 

We note that variance in DPLR form facilitates efficient 
execution of common operations such as sampling, inverse
computation and linear transformation that we detail in
the appendix.
\SILENT{
\subsection{Fast Common Operations using DPLR}\label{subsec:fast_common_operations}
We conclude this section by pointing some common operations
can be made faster when the variance is of DPLR form.
\begin{itemize}
\item \emph{Sampling}
We can combine variance propagation with sampling. To
generate $\bbz\sim N(\bbmu, \Lambda + U U^T)$, we compute
$\bbz = \Lambda^{1/2}\bbx + U\bby$ where 
$\bbx\sim N(0,I)^n$ and $\bby\sim N(0,I)^r$.

\item \emph{Inversion}
To compute Gaussian NNL and confidence regions 
we need to find $\Sigmain^{-1}$ (Or $\Sigma^{1/2}$). 
Assuming $\Lambda > 0$, inversion using the Woodbury formula \cite{GoluVanl96}
costs $O(r^3)$.

\item \emph{Multiplication}
Algorithm~\ref{algo:favour} only requires $M \bbv$, 
$M = W(\Lambda+UU^T)W^T$. We compute in $O(n^2 r)$ complexity
$M \bbv$ by $W(D(W^T\bbv)) + Y(Y^T \bbv)$
where $Y = WU$. Furthermore, using a rank $r$ approximation to $W$
leads to a sub-inference complexity of $O(n r^2)$. 
\end{itemize}
}

\section{Experiments}\label{sec:experiments}

\begin{figure}
    \centering
    \includegraphics[width=\textwidth]{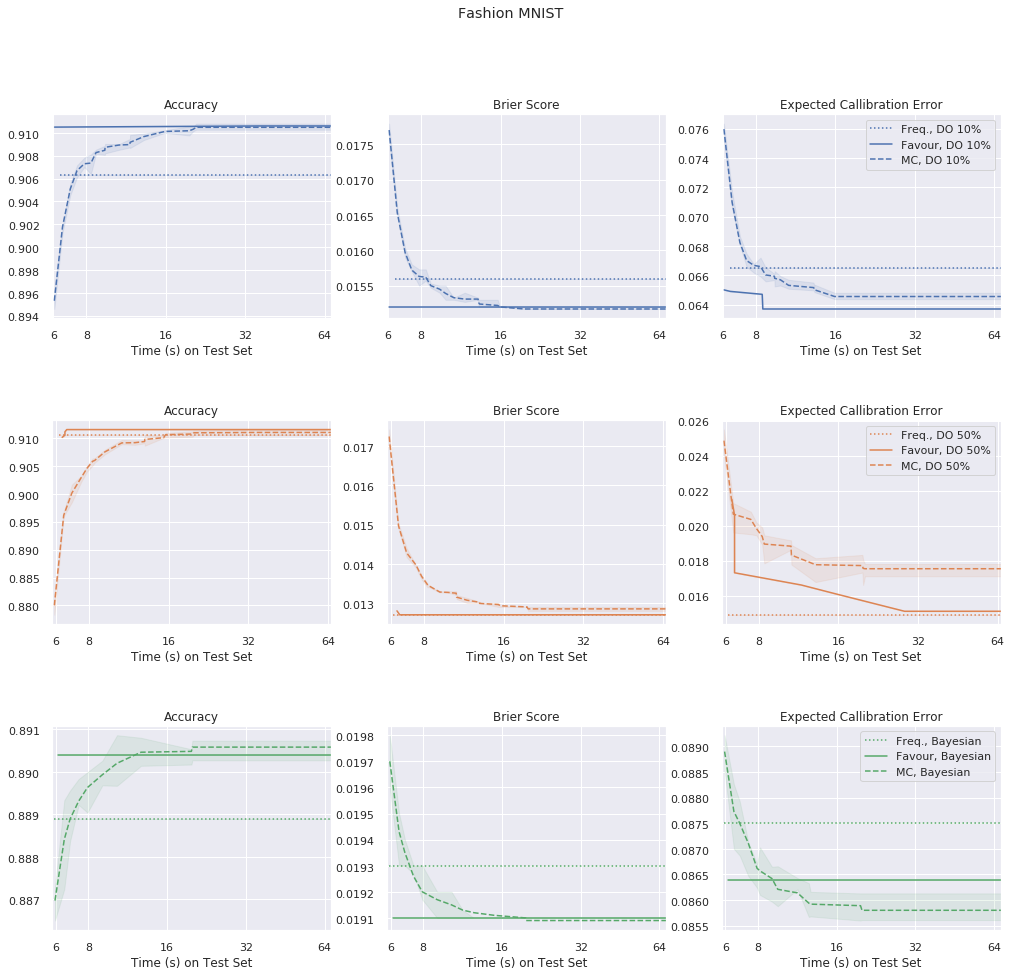}
\caption{
\textbf{Classification performance as computational budget increases.}
Favour achieves better quality of inference than MCDO when given a small computational budget. 
We trained a CNN with dropout on the top two linear layers, and computed the inference Accuracy, Brier Score and ECE for different parameter choices (e.g. number of MC samples and rank of covariance matrix in Favour.)}
\label{fig:class-calibration-fashion}
\end{figure}

\begin{table}
\centering
\hspace{-3em}
\def\arraystretch{1.5}

{\small{
\begin{tabular}{l l l | r r r r} 
 & & & \multicolumn{4}{c}{
Area Under ROC Curve
 } \\
 Datasets & Model & Inference &
 JSD & MaxP. & Ent. & Maha. \\
 \hline
 \multirow{12}{*}{\shortstack{Train dist: \\MNIST\\\\Out of dist: \\Fashion\\MNIST}}
 
  & \multirow{3}{*}{Bayes}
&Freq.  &  ---  &  $64.5 \pm 0.0$  &  $64.2 \pm 0.0$  &  $67.9 \pm 0.0$   \\
&&Favour  &  $79.6 \pm 0.0$  &  $63.6 \pm 0.0$  &  $63.3 \pm 0.0$  &  $77.2 \pm 0.0$ \\
&&MC  &  $78.5 \pm 0.1$  &  $65.3 \pm 0.1$  &  $65.0 \pm 0.1$  &  $72.2 \pm 0.1$  \\

\cline{2-7} & \multirow{3}{*}{DO 10\%}
&Freq.  &  ---  &  $77.2 \pm 0.0$  &  $77.0 \pm 0.0$  &  $93.3 \pm 0.0$  \\
&&Favour  &  $82.8 \pm 0.0$  &  $77.3 \pm 0.0$  &  $77.1 \pm 0.0$  &  $93.9 \pm 0.0$  \\
&&MC  &  $85.1 \pm 0.1$  &  $83.1 \pm 0.1$  &  $82.1 \pm 0.1$  &  $93.7 \pm 0.1$  \\

\cline{2-7} & \multirow{3}{*}{DO 25\%}
&Freq.  &  ---  &  $72.3 \pm 0.0$  &  $72.0 \pm 0.0$  &  $75.5 \pm 0.0$  \\
&&Favour  &  $88.9 \pm 0.0$  &  $72.0 \pm 0.0$  &  $71.7 \pm 0.0$  &  $78.2 \pm 0.0$  \\
&&MC  &  $90.9 \pm 0.2$  &  $73.0 \pm 0.1$  &  $72.0 \pm 0.2$  &  $79.5 \pm 0.1$   \\

\cline{2-7} & \multirow{3}{*}{DO 50\%}
&Freq.  &  ---  &  $66.6 \pm 0.0$  &  $65.7 \pm 0.0$  &  $93.4 \pm 0.0$   \\
&&Favour  &  $76.8 \pm 0.0$  &  $67.7 \pm 0.0$  &  $66.7 \pm 0.0$  &  $94.5 \pm 0.0$  \\
&&MC  &  $83.2 \pm 0.3$  &  $78.0 \pm 0.3$  &  $73.7 \pm 0.2$  &  $89.2 \pm 0.1$   \\

\end{tabular}
}
}
    \vspace{.5em}
    \caption{
    \textbf{OOD performance at similar computational costs.}
    A Le-net model was trained with dropout on the two top linear layers (alternatively Bayesian linear layers) for 400 epochs MNIST and or 250 epochs on Cifar-10.
    Each inference mode was performed on the test set of the in-distribution as well as a different distribution's test set.
    Different measures of uncertainty were computed and we report the area under the ROC curve for a decision process based on those scores.
    The inference and auc loop was performed 10 times to measure variation due to random sampling or otherwise.
    }
    \label{table:ood_auc_1}
\end{table}

To measure the time/quality trade-off of the epistemic uncertainty estimates produced by Favour, 
we evaluate it on the downstream tasks of \emph{confidence calibration} and \emph{out-of-distribution (OOD) detection}.

In \emph{confidence calibration} we desire the model to give an accurate estimate of its possible errors.
For instance, in a regression task, we expect the target value to be within the 80\% confidence interval 80\% of the time.
In a classification task, we aim for a lower calibration error \cite{ece_2017} as well as a model's ability to distinguish confident and uncertain predictions \cite{paVsPu_2018}.

In \emph{OOD detection} it is assumed that a model is more uncertain about its prediction on data points that are unlike what it has been trained on.
By transforming the uncertainty predicted by the model into a single number, we can fit a binary classifier to distinguish between uncertainty scores of in-distribution and out-of-distribution data.
If the assumption holds, and the uncertainty estimates are good, this classifier should have a large AUC-ROC.

\subsection{Confidence Calibration}

For classification tasks, we compare expected calibration error (ECE) between Favour and other sampling-based Bayesian approximations on a CNN model trained on Fashion MNIST (\cref{fig:class-calibration-fashion}) and CIFAR10 respectively. 



\subsection{Out of Distribution Detection}
For OOD detection, we use the predictive distribution to estimate the uncertainty of the model;
hence the uncertainty score can be interpreted as a discriminator score to separate in and out of distribution samples.
With this binary classification setting in mind, we use the Area Under the receiver operating characteristic curve (ROC) and precision-recall (PR) as the performance metrics.

We compute OOD based on four different uncertainty scores:
1) Jensen Shannon Divergence (JSD), which is a measure of epistemic uncertain, and 2-4) Maximum Probability, Entropy and Mahalanobis Distance, which are measures of aleoric unertainty.
(Recall the difference from \cref{fig:distribution_as_result}.)

Table \ref{table:ood_auc_1} reports our results in the case where the network is trained on MNIST and the OOD distribution is Fashion MNIST. 
We see that Favour performs similarly to MC-Dropout in this case, when given a similar computational budget.
Ideally we would like it to perform better than MCDO, at least at very small inferences times.
This will be further explored in the suplemental materials.

In the suplemental materials we consider two settings: near OOD and far OOD; where near OOD are instances where the OOD samples are from semantically similar distribution.
For near OOD, we used two settings: training on seven digits of MNIST and testing on the filtered classes; and testing a trained model on CIFAR10 on CIFAR 100.
For far OOD, we tested a trained classifier on MNIST on Fashion-MNIST, and a trained classifier on CIFAR10 on SVHN.

\SILENT{
Some ways to measure this:
\begin{description}
    \item[Max Prob]
        We take the maximum probability of the classifications.
        If the Bayesian models have high epistemic uncertainty, the sampled probabilities should even out, and the maximum should be low.
    \item[Entropy]
        We take the entropy (of the mean), which should be high when the uncertainty is high.
        Another option is to use the $\ell_2$ norm, or any higher norm, which will approximate the max.
    \item[Jensen Shannon Divergence (JSD)]
        While the two other methods are really measures of aleatoric uncertainty, this one is for Bayesian methods only.
        Only defined for Bayesian models.
        We've talked a bunch about this above.
\end{description}
}

\section{Conclusion}\label{sec:conclusion}
We have shown that variance propagation can be a viable alternative to sampling based techniques on a wide range of tasks. By our subspace iterative algorithm, variance
in DPLR form can be propagated at inference or even sub-inference complexity. While
theoretically one may need a high rank value in the ``low rank'' part of DPLR,
our experiments show that a very small rank value suffices to capture the variance
for many uncertainty rating tasks. In essence, our approach enables the actual
deployments of many BNNs. As our next steps, we plan explore how to apply this fast
variance propagation idea into BNN learning, and also explore sub-inference complexity
variance propagation of convolutional neural networks.

\SILENT{
\subsection{Future ideas}
\begin{enumerate}
    \item Could we use our approximation as a faster/alternative way to train Bayesian NNs?
    \item Could we improve the quality of the Dropout version by having trained dropout probabilities? Either per layer or per neuron?
\end{enumerate}
}


\bibliographystyle{plain}
\bibliography{refs.bib}

\appendix

\section{Appendix}

\subsection{Fast Common Operations using DPLR}\label{subsec:fast_common_operations}
The DPLR (Diagonal + Low Rank) approximation allows a number of useful operations, which we use throughout Favour:
\begin{itemize}
\item \emph{Sampling}
We can combine variance propagation with sampling. To
generate $\bbz\sim N(\bbmu, \Lambda + U U^T)$, we compute
$\bbz = \Lambda^{1/2}\bbx + U\bby$ where 
$\bbx\sim N(0,I)^n$ and $\bby\sim N(0,I)^r$.

\item \emph{Inversion}
To compute Gaussian NNL and confidence regions 
we need to find $\Sigmain^{-1}$ (Or $\Sigma^{1/2}$). 
Assuming $\Lambda > 0$, inversion using the Woodbury formula \cite{GoluVanl96}
costs $O(r^3)$.

\item \emph{Multiplication}
Algorithm \textbf{1}
only requires $M \bbv$, 
$M = W(\Lambda+UU^T)W^T$. We compute in $O(n^2 r)$ complexity
$M \bbv$ by $W(D(W^T\bbv)) + Y(Y^T \bbv)$
where $Y = WU$. Furthermore, using a rank $r$ approximation to $W$
leads to a sub-inference complexity of $O(n r^2)$. 
\end{itemize}

\subsection{Details on Mean and Variance Propagation}\label{sec:var-prop-details}

Given a random variable $\bbxin\in\R^n$ 
with mean and (co)variance $(\bbmuin, \Sigmain)$, we wish to determine the
mean and variance $(\bbmuout,\Sigmaout)$ of the output $\bbxin\in\R^m$ 
of a probabilistic neural network layer
\begin{equation}\label{eqn:general_form}
\bbxout = \act(W(D\bbxin) + \bbb).
\end{equation}
$\act(\cdot)$ is a deterministic non-linear activation function such 
as ReLU, sigmoid or tanh. The affine transformation $W\bbx + \bbb$ can 
be deterministic, that is, the entries in the $W$ matrix and $\bbb$ vector
are simple real numbers, or probabilistic when those entries are 
distributions. $D$ is a elementwise (diagonal matrix) generalized
dropout operator where each element is a scaled Bernoulli 
random variable of possibly different dropout probabilities $p$. 
That is, the value is 0 with probability $p$ and $1/q$ with probability 
$q=1-p$. The parameters $(\bbmu,\boldsymbol{\Sigma})$ characterize 
important distributions such as Gaussian that are considered good representation
of the underlying distributions of $\bbxout$.

\subsection{Propagation through Dropout Layers}\label{subsec:dropout-details}
Let $\bbp\in\R^n_+$ be the dropout probabilities for each of $\bbxin$'s components. 
$D$ is a scaled Bernoulli random variable such that $E[D] = I$, the identity matrix
yielding $\bbmuout = \bbmuin$. Next, the Law of Total Variance implies
\begin{align*}
\V[D\bbxin] &= \V[\E[D\bbxin|\bbxin]] + \E[\V[D\bbxin|\bbxin]], \\
    &= \Sigmain + \diag(\E[\bbxin^2]\hard \bbp/\bbq), \\
    &= \Sigmain + \diag((\bbmuin^2 + \diag(\Sigmain))\hard \bbp/\bbq).
\end{align*}
The computational complexity is a low $O(n)$ as it only changes the diagonal
portion of $\Sigmain$. Note that $\bbp/\bbq$ simplifies trivially to the scalar $p/q$ if dropout probabilities are identical for all of $\bbxin$'s components.

\subsection{Propagation through Linear Layers}
Consider now a linear layer $\bbxout = W \bbxin + \bbb$ where
$W \in \R^{m\times n}$, $\bbb\in\R^m$. Both $W$ and $\bbb$ can be 
distributions instead of deterministic, with $\Wmean$ and $\bbbmean$
be their respective means. Furthermore, the entries within $W$ or 
$\bbb$ need not be mutually independent while we always assume $W$ and 
$\bbb$ are independent of each other. We propagate the mean
$\bbmuout$ of $\bbxout$ by 
$\bbmuout = \Wmean\bbmuin + \bbbmean$. Let us focus on 
$\V[\bbxout] = \V[W\bbxin + \bbb] = 
\V[W\bbxin] + \Sigma_{\bbb} \in \R^{m\times m}$. The non-trivial
term is $\V[\bbz]$, $\bbz = W \bbxin$.

Let $\bbw_i$ be the $i$-th row of $W$ and consider the flattened $W$ given by
$\bbw \bydef [\bbw_1, \bbw_2, \cdots, \bbw_m]^T \in \R^{mn}$. Consider the
Kronecker product $I_m \kprod \bbx^T \in \R^{m\times mn}$. Think of it as the identity
matrix $I_m$ with each 0 replaced by a row of $n$ zeros and each 1 replaced by $\bbx^T$.
With this, $\bbxout = W\bbxin = (I_m \kprod \bbxin^T)\,\bbw$ is expressed 
as a transform on the $mn$ random variables of $W_{ij}$. 
We apply the Law of Total Variance as follows.
\[
\V[\bbz] = \V[\E[\bbz|\bbxin]] + \E[\V[\bbz|\bbxin]] 
= \Wmean \Sigmain\Wmean^T + \E[\V[\bbz|\bbxin]].
\]
For the second term, consider 
\[
    A = \V[\bbz|\bbxin] 
    = \V[(I_m\kprod\bbxin^T) \bbw | \bbxin] 
    = (I_m\kprod \bbxin^T) \Sigma_\bbw (I_m \kprod \bbx)
\]
because $(A\kprod B)^T = A^T \kprod B^T$. Note that
$\Sigma_\bbw \in \R^{mn\times mn}$ and $A\in \R^{m\times m}$. 
The $ij$-th entry $A_{ij}$ of $A$ is given by 
\[
A_{ij} = \bbxin^T \Sigma_{\bbw,ij} \bbxin 
 = \sum_{r,s} (\Sigma_{\bbw,ij})_{rs} x_r x_s .
\]
where $\Sigma_{\bbw,ij}\in\R^{n\times n}$ is the covariance 
between  $\bbw_i$ and $\bbw_j$, the $i$-th and $j$-th row of $W$.
Because $\E[x_r x_s] = cov(x_r, x_s) + \mu_r \mu_s$ we have
\[
\E[A_{ij}] = \hbox{sum all elements of}\;(\Sigma_{\bbw,ij} \hard (\Sigmain + \bbmuin \bbmuin^T)).
\]
Combining the above, we have
\begin{equation}
(\V[W\bbxin + \bbb])_{ij} = (\Wmean\Sigmain\Wmean^T + \Sigma_{\bbb})_{ij} + 
{\rm SUM}(\Sigma_{\bbw,ij} \hard (\Sigmain + \bbmuin \bbmuin^T))
\end{equation}

Equation~\ref{eqn:linear_general} simplifies in some common special cases.
\begin{enumerate}
    \item If the linear layer is deterministic, then $\Sigmain_{\bbw},\Sigmain_{\bbb}$ are zero and the variance of $\bbxout$ is simply $W\Sigmain W^T$.
    \item If the entries of $W$ and $\bbb$ are all mutually independent, such as
    in the case of mean-field VI, then $\Sigmain_{\bbb}, \Sigmain_{\bbw}$ are diagonal
    matrices. A natural way to represent $\Sigmain_{\bbw}$
    is by $W_{\sigma^2} \in \R^{m\times n}$ whose
    entries are the variances of $w_{ij}$ and for $\Sigmain_{\bbb}$ 
    the vector $\bbb_{\sigma^2}\in\R^m$. The propagated variance
    becomes
    \[
    \Wmean \Sigmain \Wmean^T + \diag\left(
    \bbb_{\sigma^2} +
    W_{\sigma^2}\,(\diag(\Sigmain) + \bbmuin^2))
    \right)
    \]
    \item Since each row of $W$ transforms $\bbxin$ to a new feature, we consider
    also the generalized mean-field VI where each different rows of the $W$ matrix
    are independent, but not so withing each row. In this case, 
    ${\rm SUM}(\Sigma_{\bbw,ij} \hard (\Sigmain + \bbmuin \bbmuin^T))$ is non-zero
    only when $i=j$ and thus only updates the diagonal of $\Sigmain$. 
    And the computation of that expression is
    \begin{equation}\label{eqn:W_independent_rows}
        {\rm SUM}(\Sigma_{\bbw,ii} \hard (\Sigmain + \bbmuin \bbmuin^T)).
    \end{equation}
\end{enumerate}

\subsection{Propagation through Univariate Activations}
An activation function $\act(\cdot)$ such as ReLU or sigmoid are univariate nonlinear functions
applied to each component of an in $\bbxin =[x_1, x_2, \ldots, x_n]^T$. If we assume Gaussian
distribution of the random variable $\bbxin$ described by $(\bbmuin, \Sigmain)$, then the mean $\bbmuout$ is given by
\[
\mu_i' = \frac{1}{\sqrt{2\pi}\sigma_i}\,
\int_{-\infty}^{\infty} \act(x) e^{-(x-\mu_i)^2/2\sigma_i^2}\;dx.
\]
For the common case of the ReLU activation, the integral has a closed form formula.
In general, numerical quadrature can be used. Regardless, $\bbmuout$ can be computed in $O(n)$ complexity. 

There are two choices for variance propagation. The first approach uses the
first order Taylor expansion at the input mean $\bbmuin$:
For $\bbxout = A(\bbxin)$, 
\[
\bbxout_i = \act(\bbmuin_i + (\bbxin_i-\bbmuin_i)) 
\approx \act(\bbmuin_i) + \act'(\bbmuin_i)\cdot (\bbxin_i-\bbmuin_i).
\]
Thus,
\[
\bbxout \approx \act(\bbmuin) + D\cdot(\bbxin-\bbmuin), \quad
D \bydef \diag([\act'(\bbmu_i); i=1, 2,\ldots, n]).
\]
This leads to the approximation $\Sigmaout \approx D\Sigmain D^T$, which is merely
scaling $\Sigmain_{ij}$ by $\act'(\bbmuin_i)\act'(\bbmuin_j)$. In particular, for
ReLU activation, this is zeroing out the $i$-th row and column whenever $\bbmuin_i \le 0$. This sparsification may have have some desirable regularization effect.

Note however that the covariance values in $\Sigmain$ remain unchanged when not made zero. This does not match the case if we assume $\bbxin$ is Gaussian because 
the variance of $\bbxout_i$ is given by
\[
(\sigma'_i)^2 = 
\frac{1}{\sqrt{2\pi}\sigma_i}
\int_{-\infty}^\infty (\act(x)-\bbmuin_i)^2 
e^{-(x-\bbmuin_i)^2/2\sigma_i^2}\;dx
\]
This motives our second approach. We define
$D \bydef \diag([\sigma'_i/\sigma_i \text{if $\sigma_i > 0$ else 0}]$, and scale $\Sigmain$ by
$D\Sigmain D^T$. This guarantees that the diagonal value
of $\Sigmaout$ matches the Gaussian assumption.

\subsection{Full Statement and Proof of JSD Approximation}

\begin{lemma}\label{lemma:jsd-details}
Let $\SUP{\bbx}{1},\SUP{\bbx}{2},\ldots,\SUP{\bbx}{n} \in \R^d$
be independent random variables with a common mean $\bbmu = \E[\SUP{\bbx}{\ell}]$ for all $\ell$ and denote
by $\bbSigma$ the variance $\E[XX^T]-\bbmu\bbmu^T$ where
$X=[\SUP{\bbx}{1},\SUP{\bbx}{2},\ldots,\SUP{\bbx}{n}]$.
Let  $\bbp\bydef \sm(\bbmu)$,
$\SUP{\bbp}{\ell} \bydef \sm(\SUP{\bbx}{\ell})$
and $J$ be the Jacobian of $\sm$
at $\bbmu$.
Then suppose the error
$\SUP{\bbDelta}{\ell} =
\sm(\SUP{\bbx}{\ell}) - (\bbp + J (\SUP{\bbx}{\ell}))$
is bounded in expectation
$\|\E[\SUP{\bbDelta}{\ell}]\| < \eps^2$ for some $\eps > 0$ where
then we can estimate the Jensen Shannon Divergence by
\[
\jsd(\SUP{\bbp}{1},\SUP{\bbp}{2},\ldots,
\SUP{\bbp}{n}) =
\left(1-\frac{1}{n}\right)\frac{1}{2}
\left( \inp{\bbp}{\diag\bbSigma} - 
\inp{\bbp}{\bbSigma\bbp} 
\right) + O(\eps^3).
\]
In particular, if $\bbSigma = \Lambda + U U^T$, this estimate is
$
\left(1-\frac{1}{n}\right)\frac{1}{2}
\left( \inp{\bbp}{\Lambda} - 
\inp{\bbp}{(\Lambda + UU^T)\bbp} 
\right)
$.
\end{lemma}

\begin{proof}
Define $\SUP{\bbdelta}{\ell}$ such that $\sm(\SUP{\bbx}{\ell}) = \bbp + \SUP{\bbdelta}{\ell}$.
The quantity of interest is
\[
Q = \jsd(\bbp+\SUP{\bbdelta}{1},\bbp+\SUP{\bbdelta}{2},\ldots,
\bbp+\SUP{\bbdelta}{n}) =
\Ent(\bbp + \bar{\bbdelta}) - \frac{1}{n}\sum_\ell\Ent(
\bbp + \SUP{\bbdelta}{\ell})
\]
where $\Ent(\bbp) = -\inp{\bbp}{\log(\bbp)}$ is the entropy.
Let $\bbg$ and $H$ be respectively the gradient and Hessian of $\Ent$ at $\bbp$.
Then by Taylor's theorem
\[
    Q = \left(\Ent(\bbp) + \inp{\bbg}{\bar{\bbdelta}} +
    \frac{1}{2}\inp{\bar{\bbdelta}}{H\bar{\bbdelta}} \right) - 
     \left(
    \Ent(\bbp) + \frac{1}{n}\sum_\ell\inp{\bbg}{\SUP{\bbdelta}{\ell}} + \frac{1}{2n}\sum_\ell
    \inp{\SUP{\bbdelta}{\ell}}{H\SUP{\bbdelta}{\ell}} 
    \right)
    + O(\delta^3),
\]
where $\delta \bydef \max_\ell \|\SUP{\bbdelta}{\ell}\|$.
Thus canceling the linear terms
\[
Q = \left(1-\frac{1}{n}\right)\frac{1}{2n}
\sum_\ell\inp{\SUP{\bbdelta}{\ell}}{-H\SUP{\bbdelta}{\ell}} +
\frac{1}{2n^2}\sum_{\ell\neq m}
\inp{\SUP{\bbdelta}{\ell}}
{H\SUP{\bbdelta}{m}}
 + O(\delta^3).
\]
By definition $\SUP{\bbdelta}{\ell} = 
J(\SUP{\bbx}{\ell}) + \SUP{\bbDelta}{\ell}$
\[
\E[\sum_\ell 
\inp{\SUP{\bbdelta}{\ell}}
{H\SUP{\bbdelta}{\ell}}]
=
\E[\sum_\ell
\inp{J (\SUP{\bbx}{\ell}-\bbmu)}
{H J (\SUP{\bbx}{\ell}-\bbmu)} ]
+ O(\eps^3)
\]
and
\[
\E[\sum_{\ell\neq m} 
\inp{\SUP{\bbdelta}{\ell}}
{H\SUP{\bbdelta}{m}}
]
=
\E[ 
\sum_{\ell\neq m}
\inp{\SUP{\bbDelta}{\ell}} 
{H \SUP{\bbDelta}{m}}]
= n^2 O(\eps^4)
\]
due to independence and $\E[ \SUP{\bbx}{\ell}-\bbmu] = 0$.
Finally,
\[
\frac{1}{n}
\sum_\ell \E[ 
\inp{J(\SUP{\bbx}{\ell}-\bbmu)}
{H J(\SUP{\bbx}{\ell}-\bbmu) } ]
= \tr(J^T H J \bbSigma).
\]
Note that $J = \diag(\bbp) - \bbp\bbp^T$ and
$H = \diag(1/\bbp)$ by
simple differentiation. Consequently
$J^T(-H)J = \diag(\bbp)-\bbp\bbp^T$.
Thus
$\tr((\diag(\bbp)-\bbp\bbp^T)\bbSigma) = 
\inp{\bbp}{\diag{\bbSigma}} -
\inp{\bbp}{\bbSigma\bbp}$. Therefore
\[
\E[ Q] = 
(1-\frac{1}{n})\frac{1}{2}\left(
\inp{\bbp}{\diag{\bbSigma}} -
\inp{\bbp}{\bbSigma\bbp}
\right) + O(\eps^3)
\]
as desired.
In general the quality of the estimate increases as $n$ grows, so we let $n\to\infty$ and get rid of the factor $1-1/n$.
This is similar to the classical Bessel's correction for covariance.
\end{proof}

\section{Additional Experiments}\label{subsec:more_experiments}

\subsection{Calibration Experiments}

\paragraph{Methodology}

We trained LeNet with 10\% dropout on the top two linear layers, on respectively MNIST, Fashion-MNIST and CIFAR 10.
The models were trained for 1000 epochs each with Adam, learning rate $2e-4$.

We ran a simple Frequentist inference, as well as using Monte Carlo samples and Favour, and compared the Accuracy as well as calibration scores.
Each method was run with a number of different parameter settings.
For sampling we used 1-1000 samples, and for Favour we used different values of variance rank, weight rank, type of ReLU propagation, and the number of subspace iterations to perform.
Inference at each parameter setting was repeated 10 times to estimate the variance in caused by random sampling.

The pareto-optimal points on the quality / time curve were identified and plot in the figures below.

The ECE and Coverage @ 95\% numbers for the sample based method is based on fitting a two dimensional Gaussian to the output point cloud.

\begin{figure}[h]
    \centering \includegraphics[width=\textwidth]{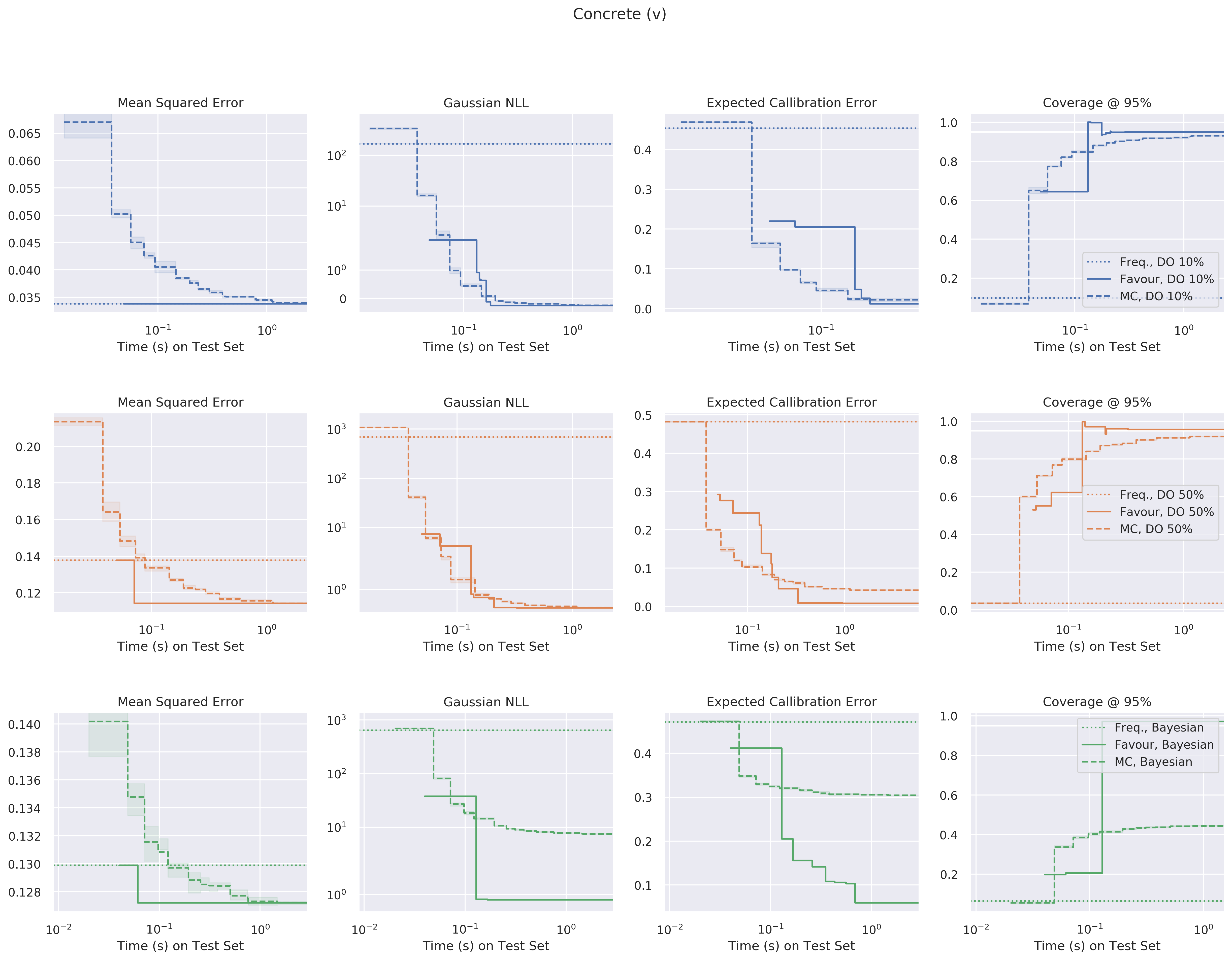}
\caption{\textbf{Regression performance on UCI/Concrete.}
We predict the strength of concrete based on 8 input variables.
Favour matches roughly the quality of MCDO estimates, when the later is given 1000 samples on the dropout models.
For the VI model however, it seems that many more samples are needed to obtain good calibration error.
}
\label{fig:reg-cali-concrete}
\end{figure}

\begin{figure}[h]
    \centering \includegraphics[width=\textwidth]{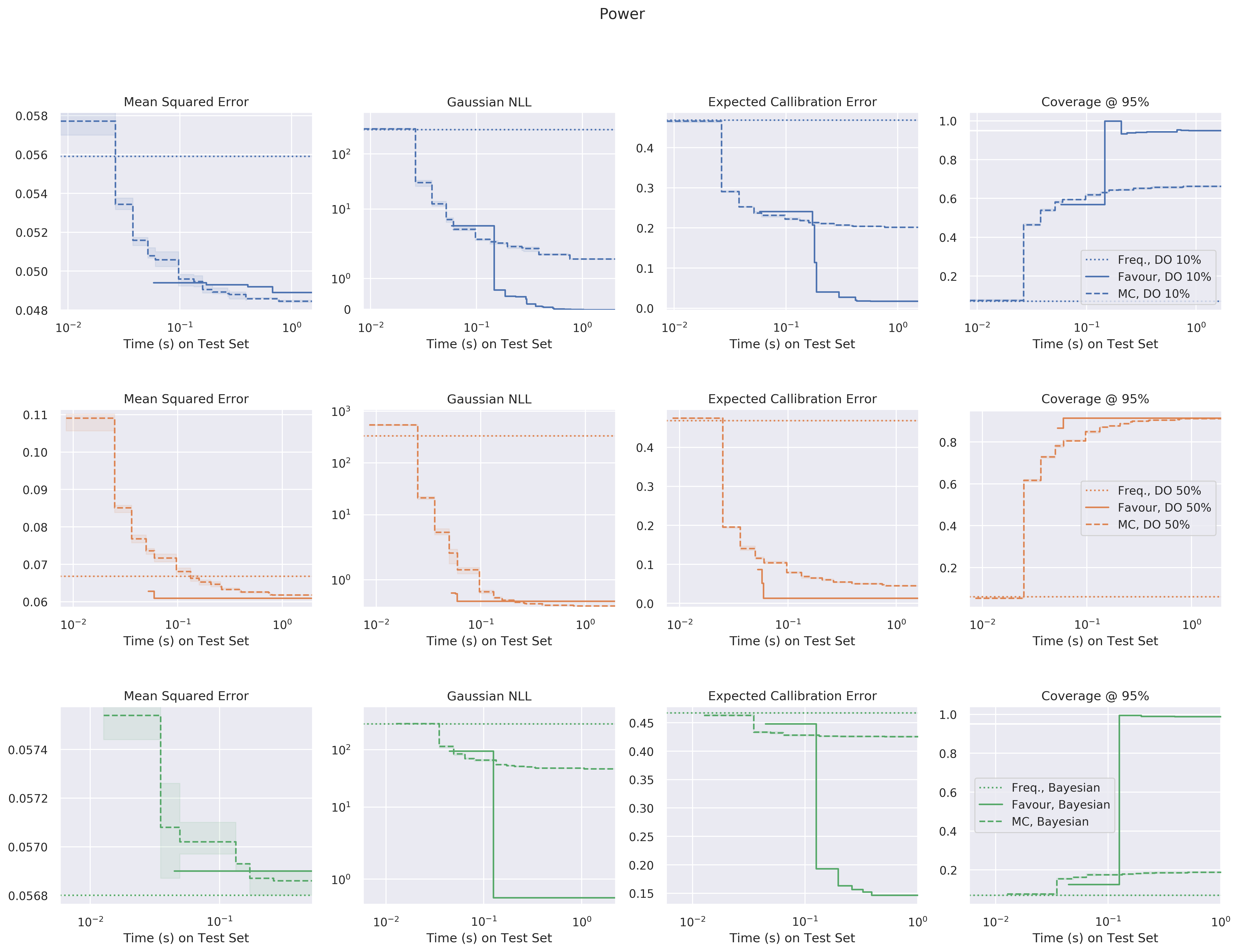}
\caption{\textbf{Regression performance on UCI/Power.}
The Combined Cycle Power Plant Data Set consist of hourly average ambient variables Temperature, Ambient Pressure, Relative Humidity and Exhaust Vacuum to predict the net hourly electrical energy output of a plant.
In the case of the 50\% dropout model, using a large amount of samples is able to outperform Favour on Gaussian NLL, but Favour uniformely achieves better ECE.
Like with Concrete, Favour is able to much better utilize the VI model than sampling.
}
\label{fig:reg-cali-power}
\end{figure}

\begin{figure}[h]
    \centering
    \includegraphics[width=\textwidth]{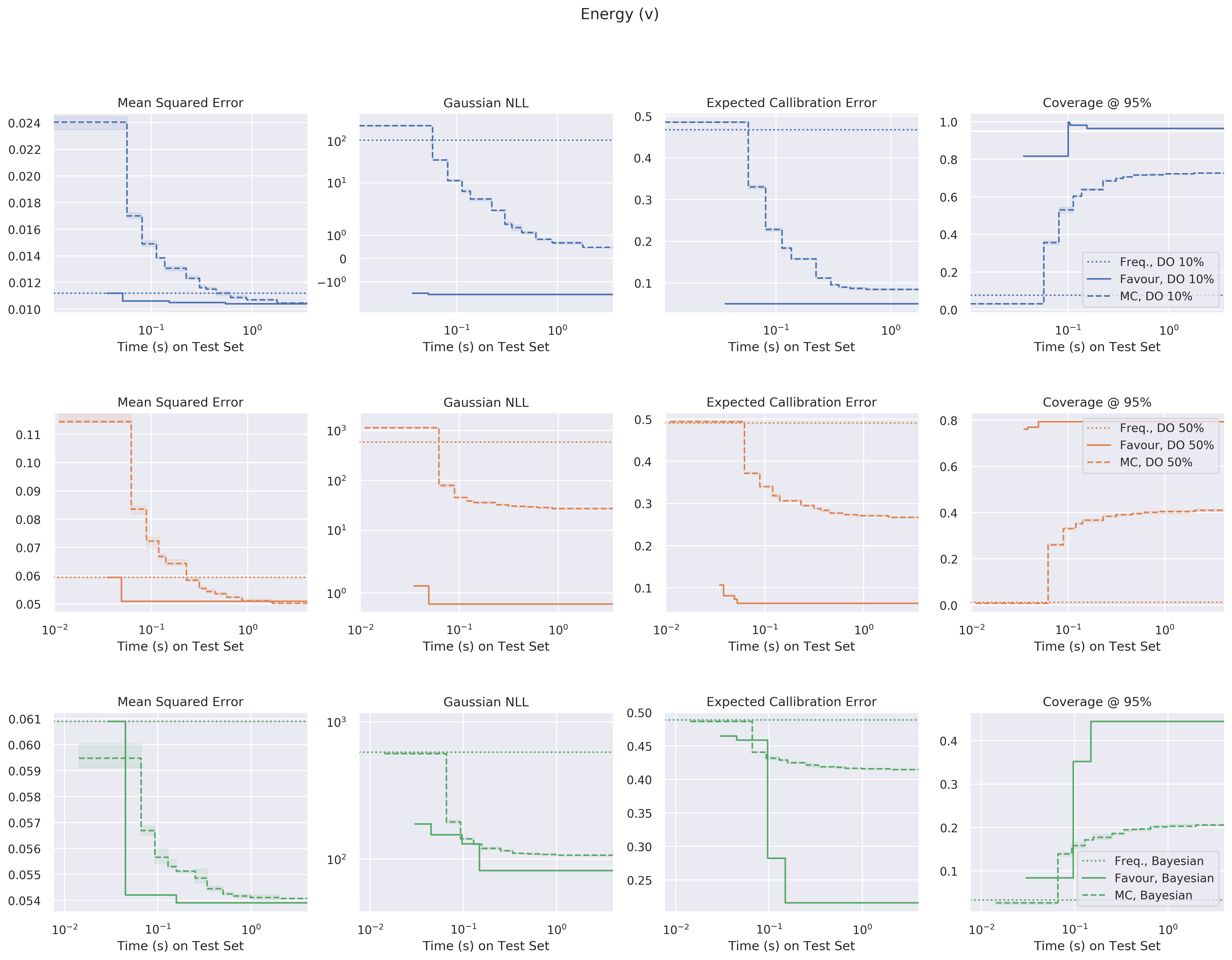}
\caption{
\textbf{Regression performance on UCI/Energy.}
We perform energy analysis using 12 different building shapes simulated in Ecotect. The test-set is not very large, which make the calibration scores unreliable. Hence we report scores on a validation set instead.
The findings are roughly the same as in the previous two experiments.
}
\label{fig:reg-cali-energy}
\end{figure}

\begin{figure}[h]
    \centering \includegraphics[width=\textwidth]{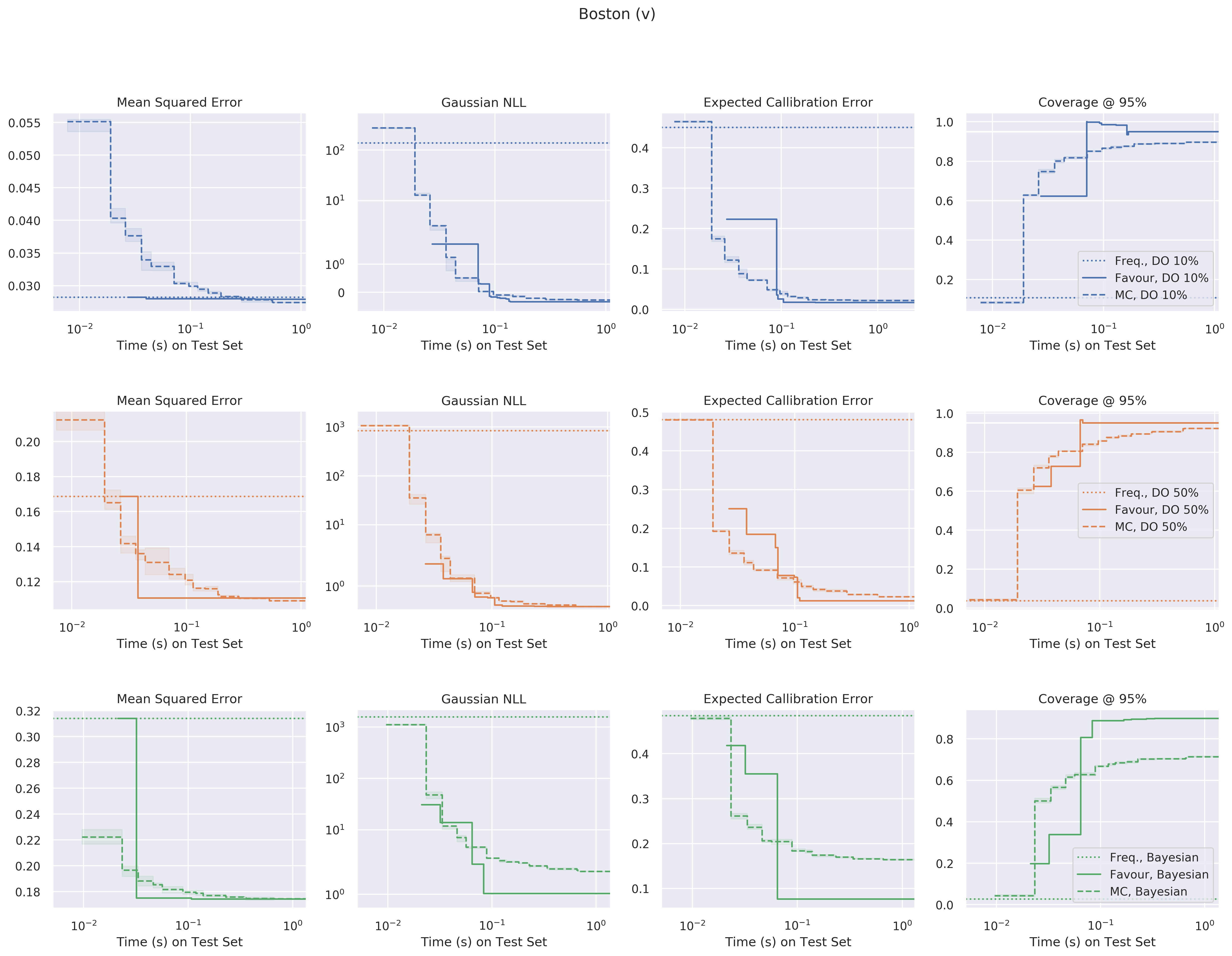}
\caption{\textbf{Regression performance on UCI/Boston}
Favour beats sample based methods at various calibration tasks, when given a medium computational budget.
For the very short time-frames sampling performs better, but then we don't win much over simple frequentist inference.
We again used a validation set which is somewhat larger than the small standard test-set, so we can get more precise calibration numbers.
}
\label{fig:reg-cali-boston}
\end{figure}

\begin{figure}[h]
    \centering
    \includegraphics[width=\textwidth]{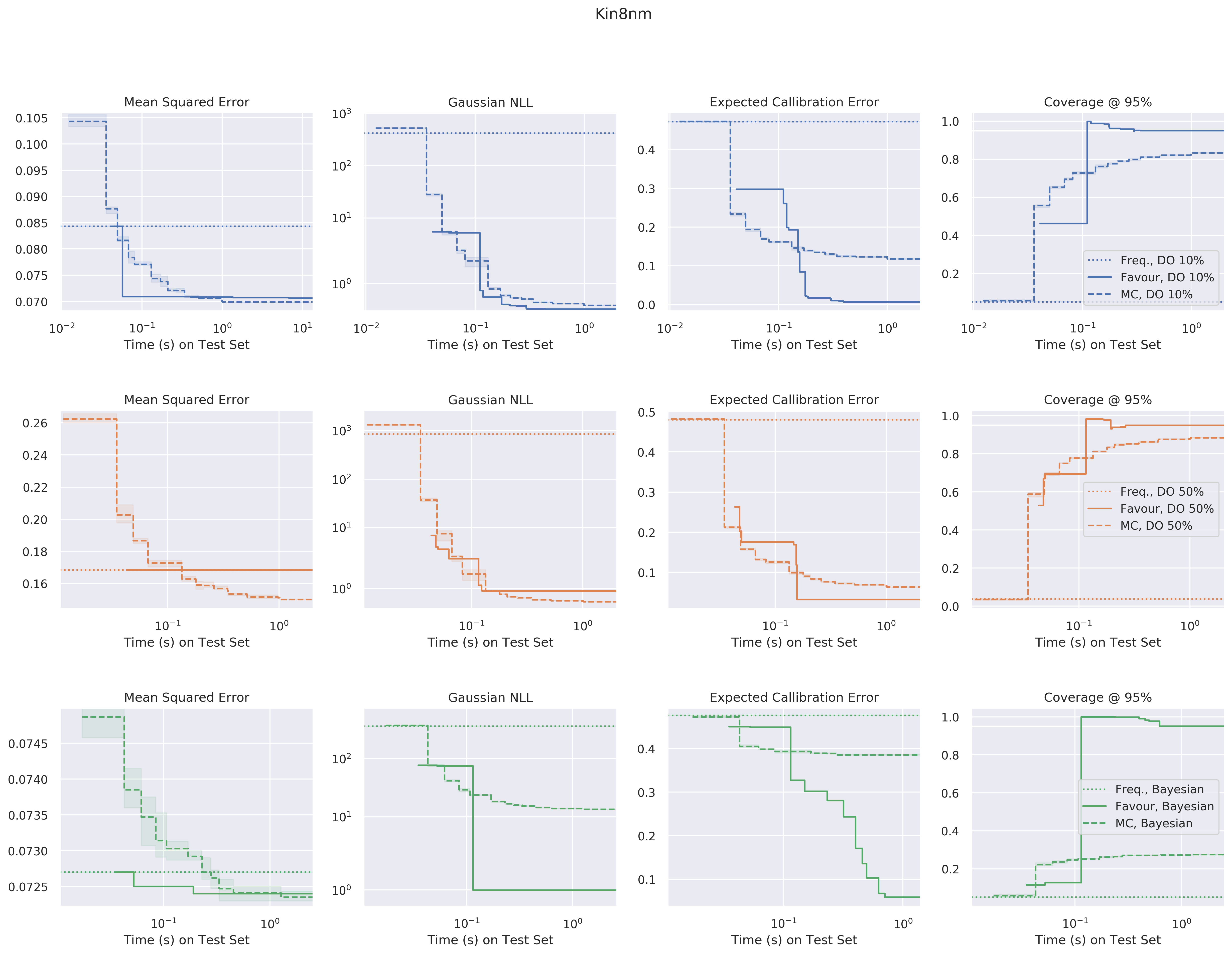}
\caption{
\textbf{Regression performance on UCI/Kin8nm.}
This is data set is concerned with the forward kinematics of an 8 link robot arm. Among the existing variants of this data set we have used the variant 8nm, which is known to be highly non-linear and medium noisy.
Favour does well on the Dropout 10\% and Bayesian models, but not so well on the Dropout 50\% model, which didn't learn the problem very well.}
\label{fig:reg-cali-kin8nm}
\end{figure}

\begin{figure}[h]
    \centering
    \includegraphics[width=\textwidth]{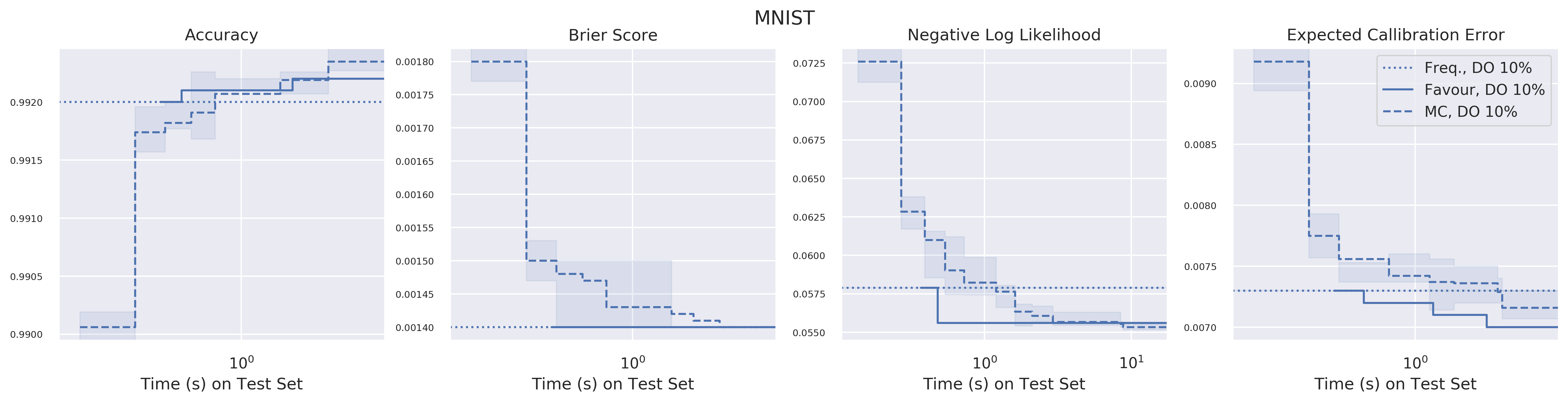}
    \includegraphics[width=\textwidth]{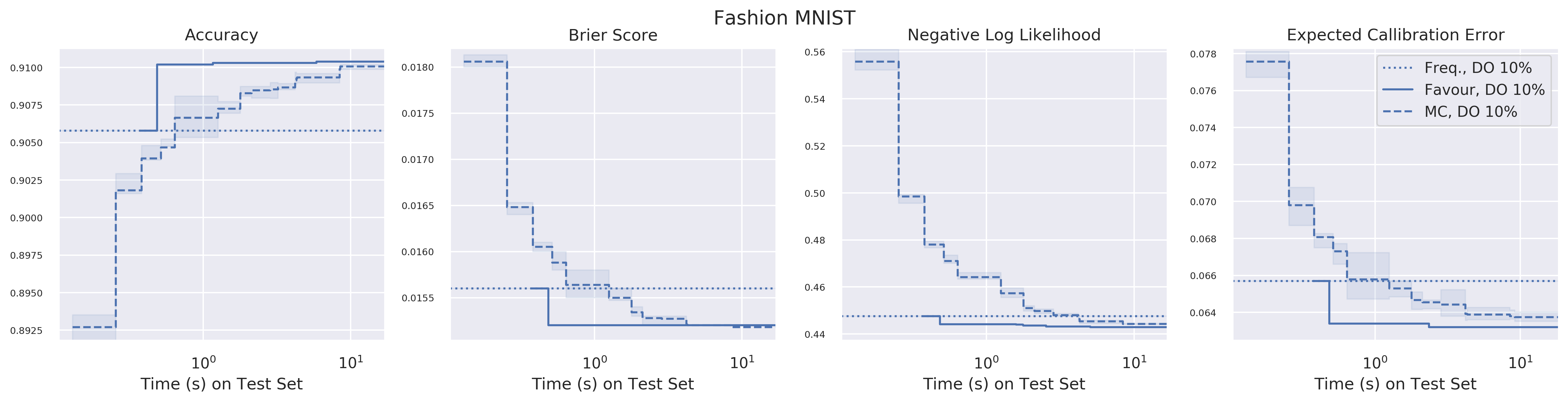}
    \includegraphics[width=\textwidth]{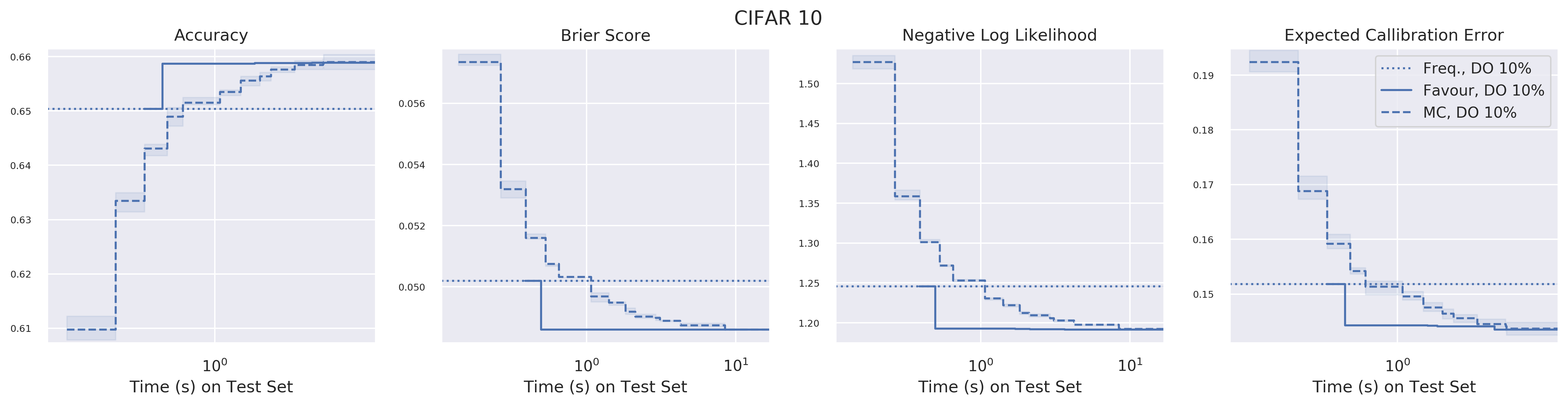}
\caption{
\textbf{Classification performance per time.}
We trained LeNet with 10\% dropout on the top two linear layers, on respectively MNIST, Fashion-MNIST and CIFAR 10.
In every case using enough samples eventually gave better results than Favour, but Favour was able to get equivalent performance results using a time budget equivalent to just 2-3 MCDO samples.
}
\label{fig:images}
\end{figure}

\FloatBarrier

\subsection{OOD Experiments}

\paragraph{Methodology}

We trained LeNet with different levels of dropout (10\%, 25\% and 50\%) as well as VI on the top two layers on repectively MNIST and CIFAR-10.
The MNIST model was trained for 400 epochs and the CIFAR-10 for 250 epochs; each with Adam, learning rate $2e-4$.

We ran a simple Frequentist inference, as well as using Monte Carlo samples and Favour on the test sets for those datasets, as well as a number of OOD datasets (Fashion MNIST and SVHN).

We computed three deterministic scores (Maximum Probability, Entropy and Mahalanobis distance) as well as two scores that are only defined in the Bayesian setting (Jensen-Shannon Divergence and KL-Diveregence.)
The scores were used in a simple in/out of distribution classifier and the performance was measured using Area under the curve, for ROC as well as Precision Recall

\begin{table}[h]
\hspace{-3em}
\def\arraystretch{1.5}

{\small{
\begin{tabular}{l l l | r r r r r} 
 & & & \multicolumn{5}{c}{
Area Under Receiver-Operating Characteristic (ROC) Curve
 } \\
 Datasets & Model & Inference &
 JSD & MaxP. & Ent. & Maha. & KL \\
 \hline
 \multirow{12}{*}{\shortstack{Train dist: \\MNIST 0-7\\\\Out of dist: \\MNIST 8-9}}

& \multirow{3}{*}{Bayes}
&Freq.  &  ---  &  $97.0 \pm 0.0$  &  $97.1 \pm 0.0$  &  $95.4 \pm 0.0$  &  --- \\
&&Favour  &  $96.8 \pm 0.0$  &  $97.1 \pm 0.0$  &  $97.2 \pm 0.0$  &  $96.0 \pm 0.0$  &  $96.9 \pm 0.0$ \\
&&MC  &  $95.8 \pm 0.1$  &  $96.8 \pm 0.0$  &  $97.0 \pm 0.0$  &  $93.7 \pm 0.1$  &  $88.1 \pm 3.2$ \\

\cline{2-8} & \multirow{3}{*}{DO 10\%}
&Freq.  &  ---  &  $89.1 \pm 0.0$  &  $97.1 \pm 0.0$  &  $79.6 \pm 0.0$  &  --- \\
&&Favour  &  $89.1 \pm 0.0$  &  $89.0 \pm 0.0$  &  $97.2 \pm 0.0$  &  $80.5 \pm 0.0$  &  $62.1 \pm 0.0$ \\
&&MC  &  $95.8 \pm 0.2$  &  $92.5 \pm 0.3$  &  $96.7 \pm 0.1$  &  $75.8 \pm 0.3$  &  $73.7 \pm 0.4$ \\

\cline{2-8} & \multirow{3}{*}{DO 25\%}
&Freq.  &  ---  &  $96.7 \pm 0.0$  &  $96.9 \pm 0.0$  &  $95.8 \pm 0.0$  &  --- \\
&&Favour  &  $96.9 \pm 0.0$  &  $96.8 \pm 0.0$  &  $97.0 \pm 0.0$  &  $95.9 \pm 0.0$  &  $57.7 \pm 0.0$ \\
&&MC  &  $95.7 \pm 0.1$  &  $95.9 \pm 0.1$  &  $96.0 \pm 0.1$  &  $86.9 \pm 0.2$  &  $79.1 \pm 1.3$ \\

\cline{2-8} & \multirow{3}{*}{DO 50\%}
&Freq.  &  ---  &  $95.9 \pm 0.0$  &  $96.0 \pm 0.0$  &  $90.8 \pm 0.0$  &  --- \\
&&Favour  &  $95.8 \pm 0.0$  &  $96.0 \pm 0.0$  &  $96.1 \pm 0.0$  &  $90.3 \pm 0.0$  &  $42.7 \pm 0.0$ \\
&&MC  &  $91.5 \pm 0.3$  &  $93.7 \pm 0.2$  &  $93.8 \pm 0.2$  &  $57.3 \pm 0.4$  &  $81.6 \pm 5.4$ \\

 \hline
 \hline
 \multirow{12}{*}{\shortstack{Train dist: \\MNIST\\\\Out of dist: \\Fashion\\MNIST}}

  & \multirow{3}{*}{Bayes}
&Freq.  &  ---  &  $64.5 \pm 0.0$  &  $64.2 \pm 0.0$  &  $67.9 \pm 0.0$  &  --- \\
&&Favour  &  $79.6 \pm 0.0$  &  $63.6 \pm 0.0$  &  $63.3 \pm 0.0$  &  $77.2 \pm 0.0$  &  $81.2 \pm 0.0$ \\
&&MC  &  $78.5 \pm 0.1$  &  $65.3 \pm 0.1$  &  $65.0 \pm 0.1$  &  $72.2 \pm 0.1$  &  $49.3 \pm 0.7$ \\

\cline{2-8} & \multirow{3}{*}{DO 10\%}
&Freq.  &  ---  &  $77.2 \pm 0.0$  &  $77.0 \pm 0.0$  &  $93.3 \pm 0.0$  &  --- \\
&&Favour  &  $82.8 \pm 0.0$  &  $77.3 \pm 0.0$  &  $77.1 \pm 0.0$  &  $93.9 \pm 0.0$  &  $82.5 \pm 0.0$ \\
&&MC  &  $85.1 \pm 0.1$  &  $83.1 \pm 0.1$  &  $82.1 \pm 0.1$  &  $93.7 \pm 0.1$  &  $88.2 \pm 0.5$ \\

\cline{2-8} & \multirow{3}{*}{DO 25\%}
&Freq.  &  ---  &  $72.3 \pm 0.0$  &  $72.0 \pm 0.0$  &  $75.5 \pm 0.0$  &  --- \\
&&Favour  &  $88.9 \pm 0.0$  &  $72.0 \pm 0.0$  &  $71.7 \pm 0.0$  &  $78.2 \pm 0.0$  &  $55.5 \pm 0.0$ \\
&&MC  &  $90.9 \pm 0.2$  &  $73.0 \pm 0.1$  &  $72.0 \pm 0.2$  &  $79.5 \pm 0.1$  &  $75.3 \pm 0.7$ \\

\cline{2-8} & \multirow{3}{*}{DO 50\%}
&Freq.  &  ---  &  $66.6 \pm 0.0$  &  $65.7 \pm 0.0$  &  $93.4 \pm 0.0$  &  --- \\
&&Favour  &  $76.8 \pm 0.0$  &  $67.7 \pm 0.0$  &  $66.7 \pm 0.0$  &  $94.5 \pm 0.0$  &  $86.6 \pm 0.0$ \\
&&MC  &  $83.2 \pm 0.3$  &  $78.0 \pm 0.3$  &  $73.7 \pm 0.2$  &  $89.2 \pm 0.1$  &  $79.4 \pm 0.7$ \\

 \hline
 \hline
 \multirow{12}{*}{\shortstack{Train dist: \\CIFAR-10\\\\Out of dist: \\SVHN}}
 
& \multirow{3}{*}{DO 10\%}
&Freq.  &  ---  &  $79.9 \pm 0.0$  &  $85.8 \pm 0.0$  &  $35.4 \pm 0.0$  &  ---\\
&&Favour  &  $65.2 \pm 5.9$  &  $80.2 \pm 0.1$  &  $86.0 \pm 0.1$  &  $36.3 \pm 0.1$  &  $82.6 \pm 11.5$\\
&&MC  &  $50.8 \pm 0.2$  &  $77.8 \pm 0.1$  &  $83.4 \pm 0.1$  &  $37.5 \pm 0.1$  &  $91.3 \pm 0.1$\\

\cline{2-8} & \multirow{3}{*}{DO 25\%}
&Freq.  &  ---  &  $88.1 \pm 0.0$  &  $85.2 \pm 0.0$  &  $82.5 \pm 0.0$  &  ---\\
&&Favour  &  $84.6 \pm 3.9$  &  $87.4 \pm 0.1$  &  $84.6 \pm 0.1$  &  $85.0 \pm 0.3$  &  $64.5 \pm 12.5$\\
&&MC  &  $71.5 \pm 0.2$  &  $85.3 \pm 0.1$  &  $80.8 \pm 0.1$  &  $83.0 \pm 0.1$  &  $83.6 \pm 0.1$\\

\cline{2-8} & \multirow{3}{*}{DO 50\%}
&Freq.  &  ---  &  $80.4 \pm 0.0$  &  $73.0 \pm 0.0$  &  $91.5 \pm 0.0$  &  ---\\
&&Favour  &  $78.9 \pm 4.1$  &  $78.0 \pm 0.4$  &  $71.4 \pm 0.4$  &  $91.4 \pm 0.2$  &  $53.6 \pm 9.2$\\
&&MC  &  $69.6 \pm 0.2$  &  $78.9 \pm 0.1$  &  $71.3 \pm 0.1$  &  $87.4 \pm 0.1$  &  $79.8 \pm 0.2$\\

\cline{2-8} & \multirow{3}{*}{Bayes}
&Freq.  &  ---  &  $83.0 \pm 0.0$  &  $84.0 \pm 0.0$  &  $32.7 \pm 0.0$  &  ---\\
&&Favour  &  $56.5 \pm 0.0$  &  $82.9 \pm 0.0$  &  $84.0 \pm 0.0$  &  $32.9 \pm 0.0$  &  $79.3 \pm 0.0$\\
&&MC  &  $58.4 \pm 0.2$  &  $82.2 \pm 0.0$  &  $83.4 \pm 0.0$  &  $32.9 \pm 0.0$  &  $75.1 \pm 0.1$\\

\end{tabular}
}
}
    \vspace{.5em}
    \caption{\textbf{Performance of OOD with AUC ROC}}
    \label{table:ood_auc}
\end{table}

\begin{table}[h]
\hspace{-3em}
\def\arraystretch{1.5}

{\small{
\begin{tabular}{l l l | r r r r r} 
 & & & \multicolumn{5}{c}{Area Under Precision-Recall Curve} \\
 Datasets & Model & Inference &
 JSD & MaxP. & Ent. & Maha. & KL \\
 \hline
 \multirow{12}{*}{\shortstack{Train dist: \\MNIST 0-7\\\\Out of dist: \\MNIST-8-9}}

& \multirow{3}{*}{DO 10\%}
&Freq.  &  ---  &  $41.5 \pm 0.0$  &  $48.0 \pm 0.0$  &  $99.6 \pm 0.0$  &  ---\\
&&Favour  &  $28.5 \pm 1.2$  &  $41.3 \pm 0.1$  &  $47.9 \pm 0.1$  &  $99.6 \pm 0.0$  &  $97.1 \pm 0.8$\\
&&MC  &  $43.8 \pm 0.8$  &  $41.8 \pm 0.8$  &  $44.9 \pm 0.9$  &  $99.4 \pm 0.0$  &  $99.4 \pm 0.0$\\

\cline{2-8} & \multirow{3}{*}{DO 25\%}

&Freq.  &  ---  &  $52.9 \pm 0.0$  &  $56.1 \pm 0.0$  &  $99.7 \pm 0.0$  &  ---\\
&&Favour  &  $46.0 \pm 1.9$  &  $51.8 \pm 0.7$  &  $55.7 \pm 0.2$  &  $99.7 \pm 0.0$  &  $94.9 \pm 2.1$\\
&&MC  &  $48.4 \pm 0.9$  &  $48.9 \pm 0.9$  &  $49.4 \pm 0.9$  &  $99.3 \pm 0.1$  &  $99.2 \pm 0.1$\\

\cline{2-8} & \multirow{3}{*}{DO 50\%}
&Freq.  &  ---  &  $64.5 \pm 0.0$  &  $65.3 \pm 0.0$  &  $99.7 \pm 0.0$  &  ---\\
&&Favour  &  $61.5 \pm 4.0$  &  $64.2 \pm 1.1$  &  $65.3 \pm 0.6$  &  $99.7 \pm 0.0$  &  $71.8 \pm 11.9$\\
&&MC  &  $54.5 \pm 1.1$  &  $56.1 \pm 1.2$  &  $56.6 \pm 1.0$  &  $97.1 \pm 0.1$  &  $97.9 \pm 0.3$\\

\cline{2-8} & \multirow{3}{*}{Bayes}
&Freq.  &  ---  &  $40.4 \pm 0.0$  &  $40.6 \pm 0.0$  &  $99.0 \pm 0.0$  &  ---\\
&&Favour  &  $45.1 \pm 0.0$  &  $42.6 \pm 0.0$  &  $42.8 \pm 0.0$  &  $98.9 \pm 0.0$  &  $99.2 \pm 0.0$\\
&&MC  &  $43.7 \pm 0.7$  &  $43.7 \pm 0.5$  &  $43.4 \pm 0.8$  &  $98.6 \pm 0.1$  &  $91.6 \pm 1.1$\\

 \hline
 \hline
 \multirow{12}{*}{\shortstack{Train dist: \\MNIST\\\\Out of dist: \\Fashion\\MNIST}}

  & \multirow{3}{*}{Bayes}
&Freq.  &  ---  &  $60.7 \pm 0.0$  &  $61.1 \pm 0.0$  &  $77.5 \pm 0.0$  &  ---
\\
&&Favour  &  $85.1 \pm 0.0$  &  $56.7 \pm 0.0$  &  $56.5 \pm 0.0$  &  $82.7 \pm 0.0$  &  $80.0 \pm 0.0$
\\
&&MC  &  $83.0 \pm 0.2$  &  $57.7 \pm 0.2$  &  $57.8 \pm 0.4$  &  $79.0 \pm 0.1$  &  $64.9 \pm 0.2$
\\

\cline{2-8} & \multirow{3}{*}{DO 10\%}
&Freq.  &  ---  &  $75.9 \pm 0.0$  &  $75.7 \pm 0.0$  &  $95.2 \pm 0.0$  &  ---
\\
&&Favour  &  $84.7 \pm 0.0$  &  $75.9 \pm 0.0$  &  $75.7 \pm 0.0$  &  $95.5 \pm 0.0$  &  $88.1 \pm 0.0$
\\
&&MC  &  $87.1 \pm 0.2$  &  $82.8 \pm 0.3$  &  $80.3 \pm 0.3$  &  $95.4 \pm 0.1$  &  $91.7 \pm 0.3$
\\

\cline{2-8} & \multirow{3}{*}{DO 25\%}
&Freq.  &  ---  &  $60.5 \pm 0.0$  &  $59.9 \pm 0.0$  &  $79.2 \pm 0.0$  &  ---
\\
&&Favour  &  $90.4 \pm 0.0$  &  $59.9 \pm 0.0$  &  $59.4 \pm 0.0$  &  $81.6 \pm 0.0$  &  $60.7 \pm 0.0$
\\
&&MC  &  $92.4 \pm 0.2$  &  $59.8 \pm 0.1$  &  $58.9 \pm 0.1$  &  $83.0 \pm 0.1$  &  $79.6 \pm 0.5$
\\

\cline{2-8} & \multirow{3}{*}{DO 50\%}
&Freq.  &  ---  &  $65.5 \pm 0.0$  &  $62.9 \pm 0.0$  &  $95.0 \pm 0.0$  &  ---
\\
&&Favour  &  $80.7 \pm 0.0$  &  $65.9 \pm 0.0$  &  $63.2 \pm 0.0$  &  $95.8 \pm 0.0$  &  $83.9 \pm 0.0$
\\
&&MC  &  $86.6 \pm 0.3$  &  $75.6 \pm 0.4$  &  $66.9 \pm 0.2$  &  $91.6 \pm 0.1$  &  $84.7 \pm 0.4$
\\

 \hline
 \hline
 
 \multirow{12}{*}{\shortstack{Train dist: \\CIFAR-10\\\\Out of dist: \\SVHN}}
 
& \multirow{3}{*}{DO 10\%}
&Freq.  &  ---  &  $89.1 \pm 0.0$  &  $92.3 \pm 0.0$  &  $60.5 \pm 0.0$  &  ---\\
&&Favour  &  $74.5 \pm 3.8$  &  $89.2 \pm 0.0$  &  $92.0 \pm 0.3$  &  $60.8 \pm 0.0$  &  $85.4 \pm 7.9$\\
&&MC  &  $67.0 \pm 0.1$  &  $87.3 \pm 0.3$  &  $89.4 \pm 0.1$  &  $61.2 \pm 0.1$  &  $93.4 \pm 0.1$\\

\cline{2-8} & \multirow{3}{*}{DO 25\%}

&Freq.  &  ---  &  $91.2 \pm 0.0$  &  $87.2 \pm 0.0$  &  $87.0 \pm 0.0$  &  ---\\
&&Favour  &  $88.6 \pm 2.4$  &  $90.4 \pm 0.1$  &  $86.3 \pm 0.1$  &  $88.3 \pm 0.2$  &  $76.8 \pm 7.6$\\
&&MC  &  $79.6 \pm 0.2$  &  $88.8 \pm 0.1$  &  $83.1 \pm 0.1$  &  $87.0 \pm 0.1$  &  $89.1 \pm 0.2$\\

\cline{2-8} & \multirow{3}{*}{DO 50\%}

&Freq.  &  ---  &  $85.5 \pm 0.0$  &  $78.4 \pm 0.0$  &  $93.8 \pm 0.0$  &  ---\\
&&Favour  &  $83.6 \pm 3.2$  &  $82.6 \pm 0.4$  &  $76.9 \pm 0.3$  &  $93.9 \pm 0.1$  &  $70.0 \pm 5.8$\\
&&MC  &  $79.4 \pm 0.1$  &  $84.2 \pm 0.1$  &  $76.9 \pm 0.1$  &  $91.4 \pm 0.1$  &  $87.5 \pm 0.2$\\

\cline{2-8} & \multirow{3}{*}{Bayes}

&Freq.  &  ---  &  $89.4 \pm 0.0$  &  $89.1 \pm 0.0$  &  $59.4 \pm 0.0$  &  ---\\
&&Favour  &  $68.3 \pm 0.0$  &  $89.4 \pm 0.0$  &  $89.1 \pm 0.0$  &  $59.5 \pm 0.0$  &  $83.1 \pm 0.0$\\
&&MC  &  $69.7 \pm 0.1$  &  $89.0 \pm 0.2$  &  $88.7 \pm 0.0$  &  $59.5 \pm 0.0$  &  $81.8 \pm 0.1$\\

\end{tabular}
}
}
    \vspace{.5em}
    \caption{\textbf{OOD performance using Precision Recall}}
    \label{table:ood_auc_pr}
\end{table}

\end{document}